# Multimodal Co-learning: Challenges, Applications with Datasets, Recent Advances and Future Directions

Anil Rahate[a], Rahee Walambe[a, b], Sheela Ramanna[c], Ketan Kotecha[a, b*]

[a] *Research Scholar, Symbiosis Institute of Technology (SIT), Symbiosis International (Deemed University), Pune 412115, India*
[b] *Symbiosis Centre for Applied Artificial Intelligence, Symbiosis Institute of Technology (SIT), Symbiosis International (Deemed University), Pune 412115, India*
[c] *Department of Applied Computer Science, University of Winnipeg, Winnipeg MB R3B 2E9, Canada*

## Abstract

Multimodal deep learning systems that employ multiple modalities like text, image, audio, video, etc., are showing better performance than individual modalities (i.e., unimodal) systems. Multimodal machine learning involves multiple aspects: representation, translation, alignment, fusion, and co-learning. In the current state of multimodal machine learning, the assumptions are that all modalities are present, aligned, and noiseless during training and testing time. However, in real-world tasks, typically, it is observed that one or more modalities are missing, noisy, lacking annotated data, have unreliable labels, and are scarce in training or testing, and or both. This challenge is addressed by a learning paradigm called multimodal co-learning. The modeling of a (resource-poor) modality is aided by exploiting knowledge from another (resource-rich) modality using the transfer of knowledge between modalities, including their representations and predictive models.

Co-learning being an emerging area, there are no dedicated reviews explicitly focusing on all challenges addressed by co-learning. To that end, in this work, we provide a comprehensive survey on the emerging area of multimodal co-learning that has not been explored in its entirety yet. We review implementations that overcome one or more co-learning challenges without explicitly considering them as co-learning challenges. We present the comprehensive taxonomy of multimodal co-learning based on the challenges addressed by co-learning and associated implementations. The various techniques, including the latest ones, are reviewed along with some applications and datasets. Additionally, we review techniques that appear to be similar to multimodal co-learning and are being used primarily in unimodal or multi-view learning. The distinction between them is documented. Our final goal is to discuss challenges and perspectives and the important ideas and directions for future work that we hope will benefit for the entire research community focusing on this exciting domain.

* Corresponding author.
*Email address:* anil.rahate.phd2018@sitpune.edu.in *(A. Rahate)*, rahee.walambe@scaai.siu.edu.in *(R. Walambe)*, s.ramanna@uwinnipeg.ca *(S. Ramanna)*, director@sitpune.edu.in *(K. Kotecha)*

# 1. Introduction

Modality refers to how things are experienced in terms of sensory inputs as sight, touch, hearing, smell, and taste. There are devices like cameras, haptic sensors, microphones etc. which directly correspond to human senses for the computational world. There are devices like motion sensors, keyboards, and physiological sensors that are indirect. Some researchers define modality as how information is represented and communicated between people, such as gesture, speech, written language, gaze, etc. [1]. Multimodality combines multiple modalities such as language, vision, audio, physiological signals, physical sensor signals, etc. are different forms of modalities used to understand the world around us and provide a better user experience. Multimodal data helps us describe the objects or phenomena using different aspects or viewpoints with complementary or supplementary information. Applications with a single modality have achieved significantly higher performance owing to the advances in deep learning techniques, computing infrastructure, and large datasets. As early as 2009, studies [1] have shown that using multiple modalities can improve performance over a single modality. The recent research has shown further improvements with the latest deep learning methods. Hence, there is an increased research focus on multimodal machine learning or deep learning.

### *1.1 Multimodal deep learning*

Multimodal applications provide more accuracy and robustness than single modality applications as they combine information from multiple sources at signal level or semantic level, referred as multimodal fusion. The systems which support communication with humans using different modalities are referred to as multimodal systems. Multimodality is also defined as capacity of a system to interact using multimodal communication by processing information automatically. The modalities can cooperate in six ways: equivalence, transfer, specialization, redundancy, complementarity and concurrency [2]. The study of multimodal systems started in the 1980s with the system *'put-that-there'*, which had speech and the location of a cursor as input to create contextual reference [3]. Several studies followed involving different modalities and applications: speech and gesture for conversation understanding [4], audio, video, language for emotion analysis [5], image and text for sentiment analysis [6], mental health monitoring using wearable sensors [7], RGB and depth for medical imaging [8], radar images and optical images for weather mapping [9], language translation using text and image [10], cross-media retrieval using EEG and image [11]. The fusion of gas sensors and thermal images [12], provides better accuracy and robustness than individual models for gas detection.

Multimodal machine learning taxonomy [13] provided a structured approach by classifying challenges into five core areas and sub-areas rather than just using early and late fusion classification. These five technical challenges are representation, translation, alignment, fusion, and co-learning, as shown in Fig. 2. Representation represents and summarizes multimodal data such that complementary and supplementary information is utilized in a model [14]. Alignment identifies mapping between modalities, and translation focuses on changing data from one modality to another. Fusion combines information from multiple modalities to achieve a prediction task. Co-learning, which is the focus of this paper, is about the transfer of knowledge between modalities.

Although these five challenges appear distinct, there is an overlap among them.

Multiple challenges are applied together to achieve several tasks using multimodal deep learning models. For example, we need representation to apprehend complementary and supplementary information before fusing those modalities. Likewise, representation is required for the tasks related to translation and alignment. For co-learning also, methods from all remaining four challenges are used to achieve its objectives.

### *1.2 Need to study multimodal co-learning*

The higher performance of multimodal models depends on the availability of aligned, noiseless, and annotated modalities at training and testing. However, all modalities may not be available at all times; those may be noisy and may be in a limited amount. For example, the system must simultaneously process speech and gestures with poor acoustic and visual conditions, variations in dialects, and light conditions to understand speech well. Hence, multimodal systems need the capability to deal with missing or noisy conditions. This key practical consideration is addressed by multimodal co-learning, which uses the knowledge transfer from one modality (informative) to another modality (less informative). Empirical evaluation and theoretical studies [15] have shown that multimodal co-learning provides a higher performance on sentiment analysis when trained on audio, video, and language and tested on language than the models trained and tested only on the language modality. It has exhibited adaptability to another dataset of the same domain and datasets from a different domain.

Thus, multimodal co-learning is critical to realize the potential of multimodal applications that can work in real-life-like situations. With the increasing use of sensors, cameras, physiological devices, mobile devices, medical imaging, etc., multimodal data became easily available for multimodal applications. The multimodal applications are used in multiple areas such as affective computing, industrial decision and control systems, multimedia, autonomous systems, medical systems, military equipment, satellite systems, etc. The multimodal systems in these applications need to be robust and provide the required prediction accuracy to poor signals or different conditions, to avoid life-threatening as well as catastrophic consequences.

The multimodal co-learning can be treated as a dark horse of multimodal deep learning with its enormous potential to support practical life applications. However, co-learning has not received the required focus as a separate topic, beyond initial classification [13] into three sub-areas using data parallelism, namely, parallel, non-parallel and hybrid. There are a number of studies focusing representation [14,16–18], alignment [19–22], translation [21,23], and fusion [12,24–28]. However, the co-learning challenge is discussed only in [15] . Similarly, multiple surveys have focused on the use of multimodal deep learning in a particular domain. None of these surveys covered multimodal co-learning or attempted to define taxonomy.

Hence, we conducted this research study to comprehensively examine the multimodal co-learning area, which can help the multimodal deep learning field leap ahead.

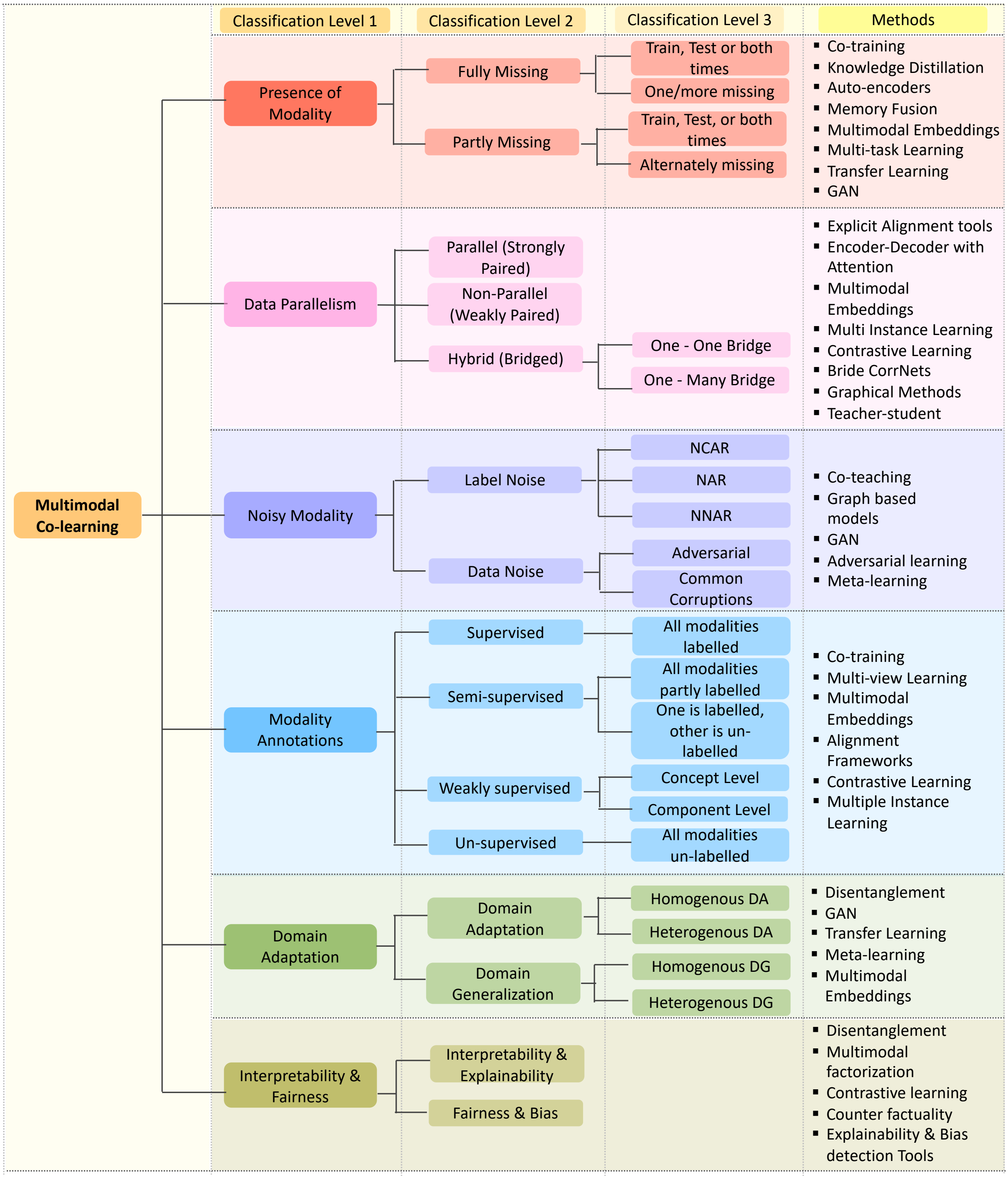

**Fig. 1.** Our proposed multimodal co-learning taxonomy

### *1.3 Contributions*

This paper endeavors to thoroughly investigate multimodal co-learning including, recent advances, challenges, datasets, and applications. We believe this is the first work to study multimodal co-learning beyond the initial co-learning taxonomy of data parallelism [13], as shown in Fig. 2. We reviewed the existing categories, identified additional categories based on recent literature, and presented the latest frameworks supporting multimodal co-learning and modality conditions during training and testing.

Also, we showed how methods from other areas, namely, representation, alignment, translation, and fusion, assist co-learning or overlap with co-learning. To bring a historical perspective, we reviewed and explained the techniques established in unimodal scenarios, used for co-learning in the early period of multimodal evolution. Finally, we presented key challenges, perspectives, datasets, applications, and important directions for future work.

The goals of this survey can be summarized as follows:

- As shown in Fig. 3, we established the co-learning objectives that can include existing co-learning work and are specific enough to clarify the purpose of co-learning.
- We expand the initial co-learning taxonomy to a comprehensive taxonomy, as shown in Fig. 1, which helps set future research directions in multimodal co-learning.
- We offer an extensive literature review and organize it as per the proposed taxonomy from the viewpoints of data, models, and applications. We also presented a summary of insights and a discussion on each objective, establishing a better understanding of co-learning.
- We propose promising future directions for co-learning in problem setup, techniques, applications, and datasets. These recommendations are based on the analysis of current constraints to achieve co-learning objectives.

### *1.4 Organization of the survey*

The rest of this paper is organized as follows: In Section 2, we set the direction by defining the objectives of multimodal co-learning and the research goals of this study. In Section 3, we explained the background with the help of single modality techniques supporting objectives of multimodal co-learning for unimodal tasks. In Section 4, we proposed comprehensive multimodal co-learning taxonomy with classification and sub-classification. We presented our investigation of research studies that implemented multimodal co-learning objectives. Section 5, shared details on deep learning methods used for multimodal co-learning implementations, including the emerging ones. In Section 6, multimodal co-learning applications and available datasets are covered. Section 7 discussed open problems and future directions in this active field, and finally, we conclude in Section 8.

## 2. Research goals and methodology

A key objective of multimodal co-learning is to work in real-life conditions where one or more modalities are scarce during training and testing. Those may be noisy, un-labeled, or with incorrect labels. We exerted these conditions as formal objectives of multimodal co-learning, as shown in Fig. 3, to guide our study of research papers in this area.

Multimodal co-learning being less investigated area to date, only a few papers discuss multimodal co-learning. However, we have observed that research studies and implementations focus on one or more multimodal co-learning objectives without explicitly referring to them as co-learning objectives. Hence, we systematically searched and studied papers that mentioned one or more multimodal co-learning objectives to answer research questions as listed in Fig. 4.

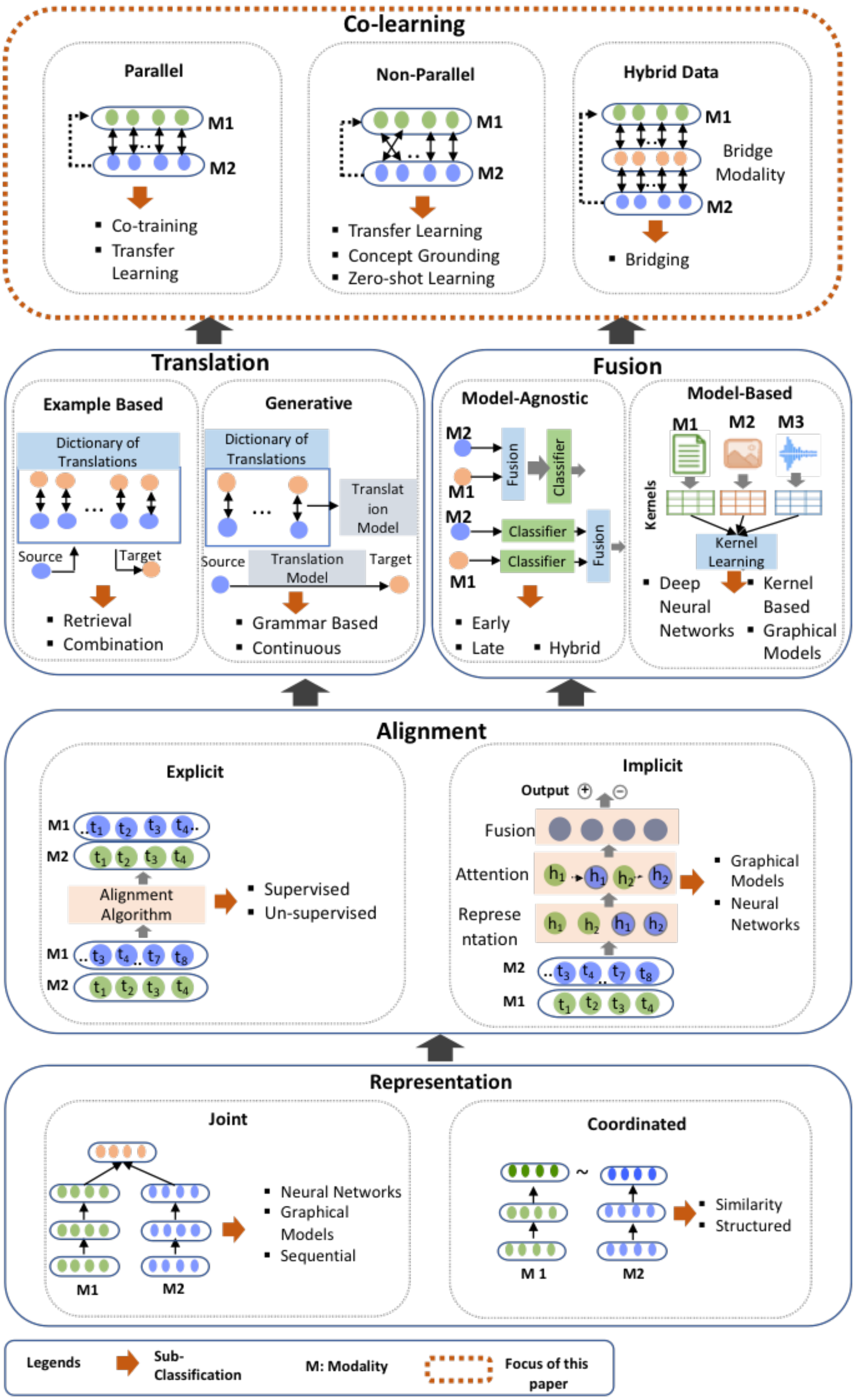

**Fig. 2.** Multimodal machine learning taxonomy based on challenges mentioned in [13])

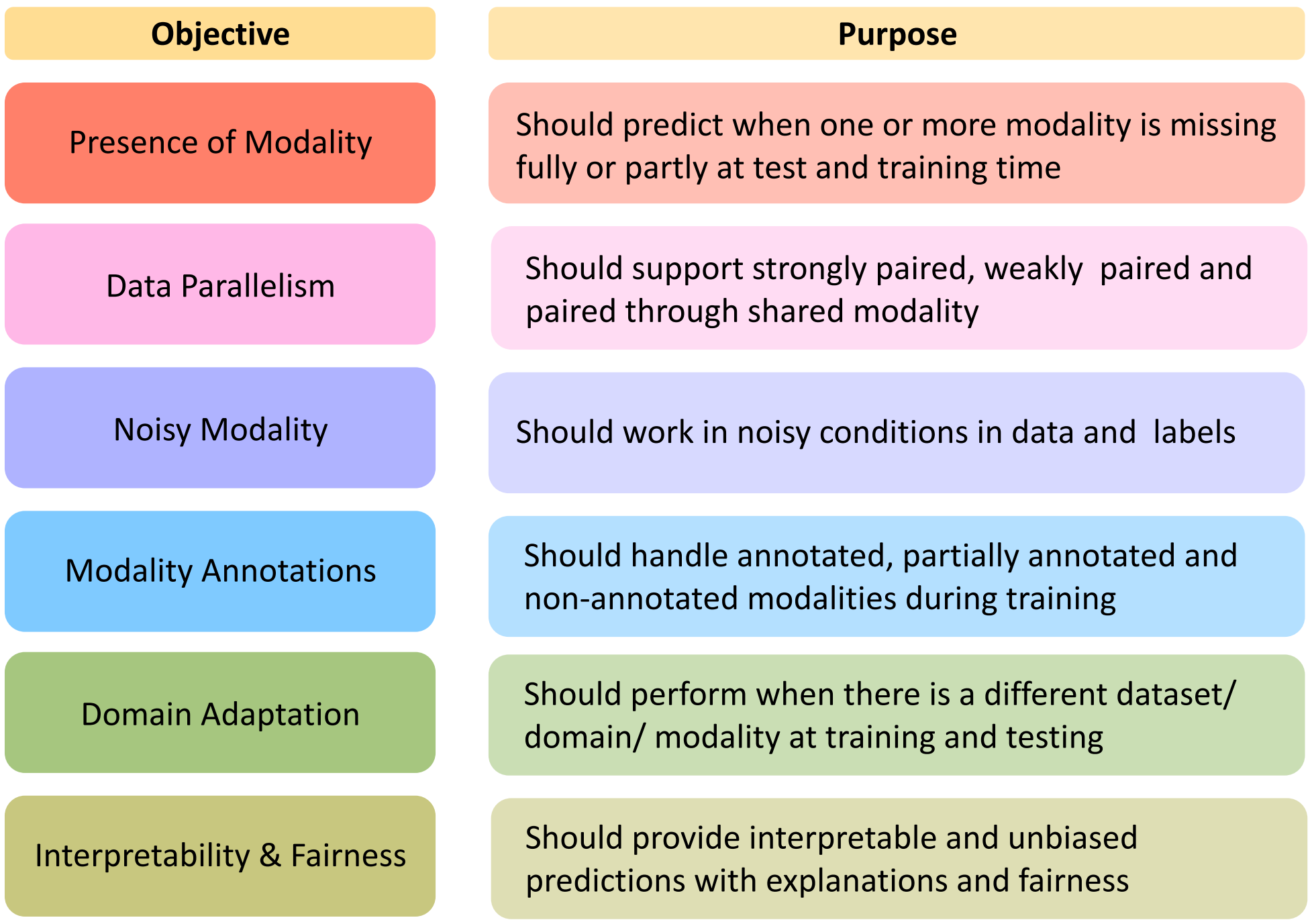

**Fig. 3.** Multimodal co-learning objectives

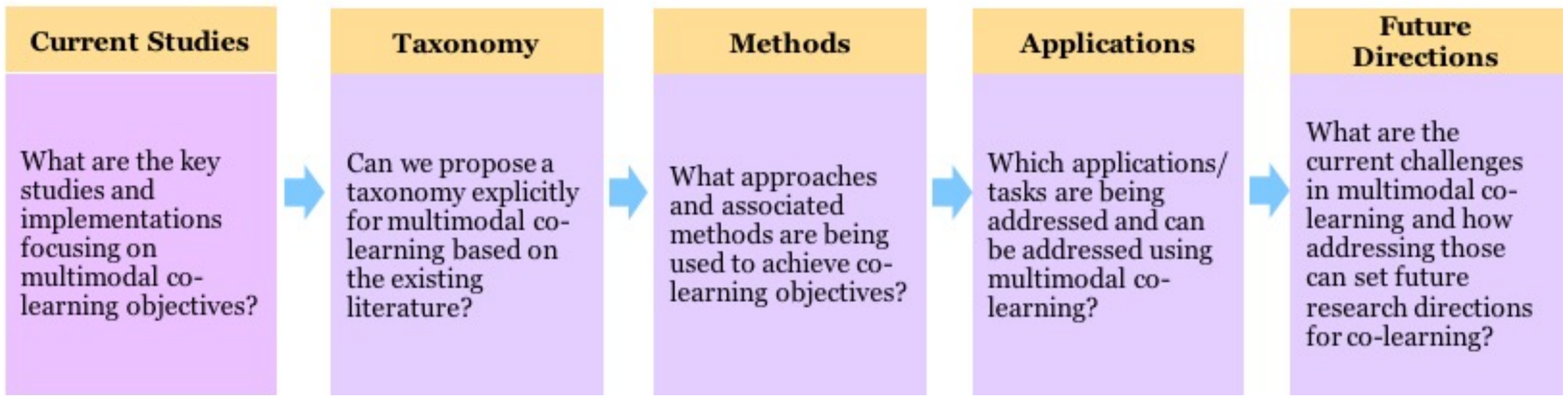

**Fig. 4.** Research questions used for this study

We also studied papers that mentioned co-learning objectives for single modality models to present comparison and historical perspective to the researchers.

## 3. Historical perspective to multimodal co-learning

Data from different sources, which can be grouped into different views, employ Multi-View Learning. Single view models were not able to handle heterogeneous data, and each view has different statistical properties. Multi-view learning introduces one function to address one view, and the functions of all views are optimized to improve the performance [19]. For example, in remote sensing, optical images and synthetic aperture radar (SAR) images are captured using two different sensors providing two different views to predict land cover mapping. Similarly, image and text modalities in image captioning can be treated as two different views for multi-view learning methods. Therefore, we briefly discuss multi-view learning in this section.

We also reviewed some prevalent techniques in unimodal space to address the co-learning objectives for single modality tasks. This discussion provides the necessary background to set the context for multimodal co-learning while conveys clarity by understanding similarities and differences. Although we have discussed these techniques in this section focusing on the historical perspective, these are still employed for multiple implementations. These techniques are getting improved in combination with the latest deep learning techniques for unimodal and multimodal tasks.

### *3.1 Co-teaching*

Noisy labels are a common occurrence in real life and specifically mean label corruption from the ground truth. It affects the performance and robustness of the models. Noisy labels are challenging for deep neural networks, as these models can memorize even the noisy labels with high accuracy. The research for handling noisy labels in deep learning was initiated by estimating the noise transition matrix to focus on finding clean labels from noisy ones and updating the network. *MentorNet* [29] supervises the Student network training by providing a sample weighting scheme to select clean instances to guide the training. *Decoupling* [30] trains two networks and updates only in case of prediction disagreement between two rather than a label; this way, the disagreement decreases when predictors get better and maintain a constant rate of noisy labels.

Lately, *Co-teaching* [31] showed better performance with high to low noisy labels (45%-20% noisy labels with multiple classes). The two networks are trained simultaneously, and each network views its small loss instances as useful in each mini-batch. However, the model allows both networks to teach others in every mini-batch allowing error to flow between them.

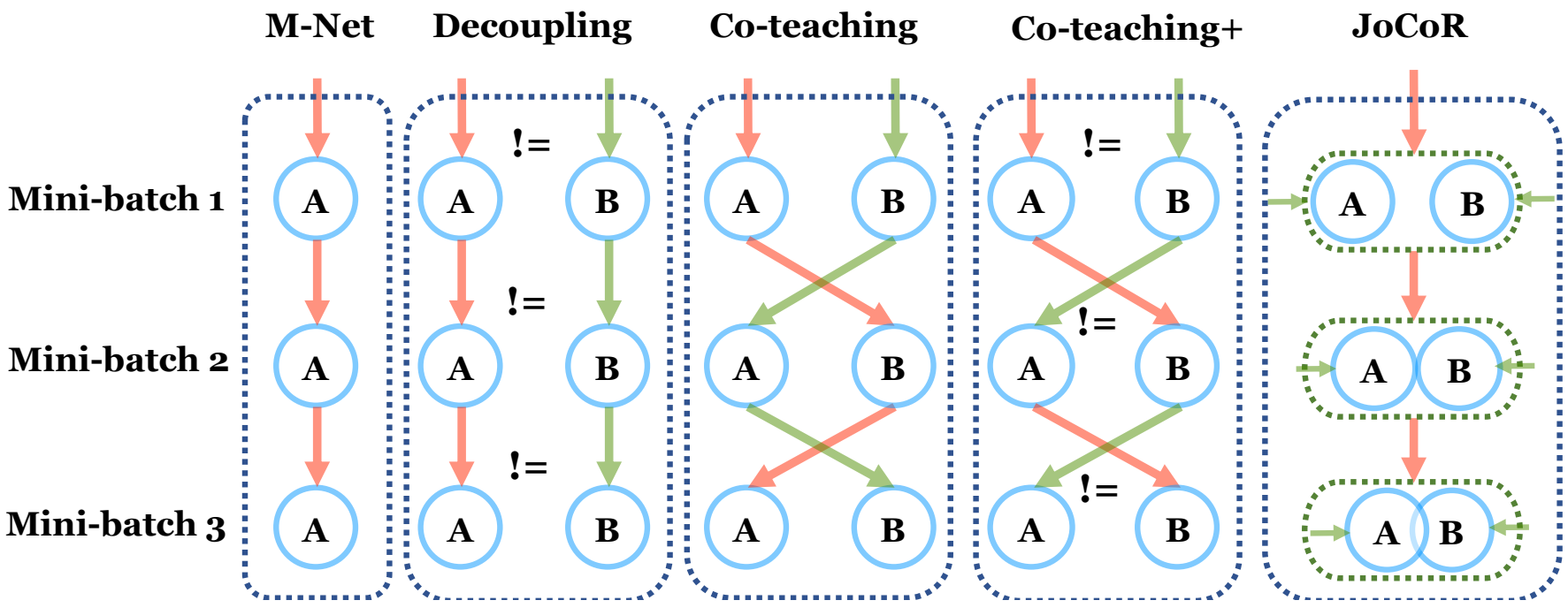

**Fig. 5.** Comparison of error flow among MentorNet (M-Net), Decoupling, Co-teaching, Co-teaching+ and JoCoR (adapted from [31], [32])

With the increase in the number of training epochs, it is observed that two networks in the co-teaching model converge to consensus and lose their advantage over MentorNet. *Co-teaching+* [33] addresses this issue by keeping two networks divergent even for a large number of epochs by modifying the 'update by disagreement' strategy by having data update (disagreement-update) step and parameters update (the cross-update) step. Another extension of co-teaching is named JoCoR (**Jo**int Training with **Co-R**egularization) [32], which trains two networks with a joint loss – supervised loss and the co-regularization loss. In Fig. 5, red and green arrows denote error flows in network A and network B assuming error comes from selecting training instances. Only one network (A) is maintained in M-Net, whereas two networks (A and B) are updated when their predictions disagree in decoupling. In co-teaching, each network considers its small loss predictions as helpful and trains its peer network. In co-teaching+, the peer network is updated in case of prediction disagreement. Co-teaching and co-teaching+ displays error flow in a zigzag shape. JoCoR also has two networks but trains them with a joint loss.

### *3.2 Teacher-Student networks*

Having a large volume of labeled data for deep learning is still a challenge; hence alternate approaches of having some labeled data and rest unlabeled data are in focus and termed as Semi-Supervised Learning (SSL). Proxy label methods generate proxy labels on unlabeled data using a prediction function. These methods are divided into two main classes: self-training and multi-view learning. In self-training, the model produces proxy labels, whereas, in multi-view learning, different data views are used to train models.

Self-training typically has a similar or higher capacity student model. However, the self-training model cannot correct itself, and any bias or error in predictions is amplified, yielding confident but wrong proxy labels on unlabeled data [34]. Therefore, it is essential to decide which pseudo labels should be appended to the labeled dataset. This process is the underlying principle for the *teacher-student* network. Initially, the teacher-student network was introduced as a *distillation* to compress a large model (teacher) into a smaller one (student); the student model is typically smaller and faster than the teacher model.

Recently, adding fine-tuning to student network on labeled images as a final step and improved data selection strategy [35] has shown the state of the art performance on image and video classification. The model utilizes 1 billion unlabeled images as a training set and the vanilla ResNet-50 [36] model. Another extension, as a robust teacher-student algorithm [37], adds robustness against perturbations (noise) to the student network. The robustness is added by modifying the objective function to minimize the difference between the gradients of student and teacher networks with Gaussian noise during training and testing. Like co-teaching, teacher-student deep semi-supervised learning (TS-DSSL) [38] also performs better with uniform and non-uniform noise distributions.

Combining teacher-student network with a pre-trained language model [39] as a data augmentation approach is used to extract task-specific in-domain data from a large bank of web sentences. This model has improved performance over the pre-trained model baseline RoBERTa [40], knowledge distillation, and few-shot learning.

### *3.3 Co-training*

Co-training is a form of multi-view learning algorithm of SSL methods and expects that two independent views represent every data point. The algorithm creates weak classifiers that utilize proxy labeling procedures to add more labeled data points using the threshold set on the prediction of classifiers. Blum and Mitchel [41] combined labeled and unlabeled data with co-training for creating more training samples for web page classification in their seminal work. They used two views – content on the web page itself and an anchor text in the hyperlink for the web page with Naïve Bayes as classifiers.

Co-training needs two different views of the same data point; however, data has only one view in many cases. In some instances, multiple views are generated by adding noise or by using data augmentation techniques. Adversarial examples are employed to create different views [42] and to stop two networks from collapsing. The obtained deep co-training model for semi-supervised image recognition showed performance improvement on Street View House Number (SVHN) [43], CIFAR 10/100 [44], and ImageNet [45] datasets. In the medical field, especially medical imaging, having labeled data is costly and time-consuming as experts must create masks or boundaries of objects. Hence, 3D volumes were rotated and permuted [8] to form multiple views for co-training for medical image segmentation of the pancreas segmentation dataset. Co-training based model is used for RGB-D object recognition [46] by having RGB and depth as two views and training two networks on these views. Convex clustering identifies different attributes for each class and adds samples from unlabeled to labeled datasets based on uniform distribution of attributes to deal with class imbalance.

Co-training is extended [47] to classify weakly labeled videos downloaded from the web using five modalities - RGB, motion, audio, concatenation of RGB-motion-audio, and metadata. Multiple Kernel Learning (MKL) is employed to classify images using images and tags as two modalities in a semi-supervised manner [48]. In this, the image-only classifier is utilized during test time with missing tag data modality.

### *3.4 Discussion*

In this section, we set the context by explaining how objectives promised by multimodal co-learning are achieved in unimodal settings and multi-view learning. Co-training is well suited for multimodal data as each modality can be considered as different views. One modality can assist other modalities during training that may not be present during testing. Teacher-student networks and other self-training methods help achieve the co-learning objectives of semi-supervised or weakly supervised annotations and deal with missing or noisy data samples. Co-teaching methods help noisy and weak label conditions, which are also co-learning objectives. This section provides information on some of the earlier methods, which started with unimodal data are extended to multimodal data.

Further details on implementation of these multimodal co-learning methods are discussed in the following sections. The recent techniques like Encoder-Decoder [49], Attention models [50], Transformers [51], Generative Adversarial Networks (GAN) [52], Zero-Shot Learning (ZSL) [53], Multi-task Learning [54], Transfer Learning [55], Meta-Learning [56], Multiple Instance Learning [57], and Domain Adaptation [58] also facilitate achievement of the multimodal co-learning objectives and are discussed in the following sections.

## 4. Multimodal co-learning taxonomy

In multimodal co-learning, one modality acts as a supporting modality and aids another modality with the transfer of knowledge between modalities during training. This supportive modality is usually not present at inference time. The same is applicable for more than two modalities. Earlier work [13] has classified multimodal co-learning based on data parallelism at training time, as shown in Fig. 2. Based on our analysis of research papers that addressed co-learning objectives, we believe it is crucial to expand the co-learning taxonomy proposed earlier in [13], making it comprehensive to encourage new research activities.

Our proposed taxonomy is shown in Fig. 1, is based on key considerations of modality conditions at training and test time. These conditions are the type of noise, the number of missing modalities, availability of annotated modality data partially or entirely, the pairing of data and data from different datasets or domains. In the proposed co-learning taxonomy, we included the data parallelism category from multimodal machine learning taxonomy. Methods in data parallelism are not included in the proposed taxonomy as methods are part of the implementation.

The subsequent sections cover each of the co-learning objectives in terms of objective definition, objective classification, recent studies and methods used, the outcome they achieved, data strategy, and applications addressed. We consider multimodal sentiment classification using three modalities, audio, video, and text, to illustrate some of the concepts.

*4.1 Presence of modality*

The foremost purpose of multimodal machine learning is to create models that utilize information from multiple modalities, to have higher accuracy than the unimodal models. The researcher's focus has helped us to arrive at state-of-the-art multimodal deep learning models. In various instances, it is assumed that all the modalities are available at training and test time. However, in real life, all the modalities may not be present at the test time and sometimes may not be present during training. This challenge is addressed by multimodal co-learning, making models robust for missing modalities and noisy data inputs. The development of models for missing modality conditions is also helping us answer a question: will the integration and fusion of multimodal information in training help even if the task is unimodal at test time? [15]. The presence of modality is a crucial factor for multimodal co-learning models. Hence, we consider it as one of the explicit classification criteria for the proposed comprehensive co-learning taxonomy.

Various modality conditions at training and testing are shown in Fig. 6, namely, a) all modalities are present at train and test time, b) all modalities are present at train time, and one modality is missing at test time, c) all modalities are present at train time, and two modalities are missing at test time, d) one modality is missing at train time, and all modalities are present at test time, e) two modalities are missing at train time, and all modalities are present at test time, f) all modalities are present at train time, and one modality is missing partly at test time, g) all modalities are present at train time, and two modalities are missing partly at test time, and h) two modalities are missing partly at train time and two modalities are missing partly at test time.

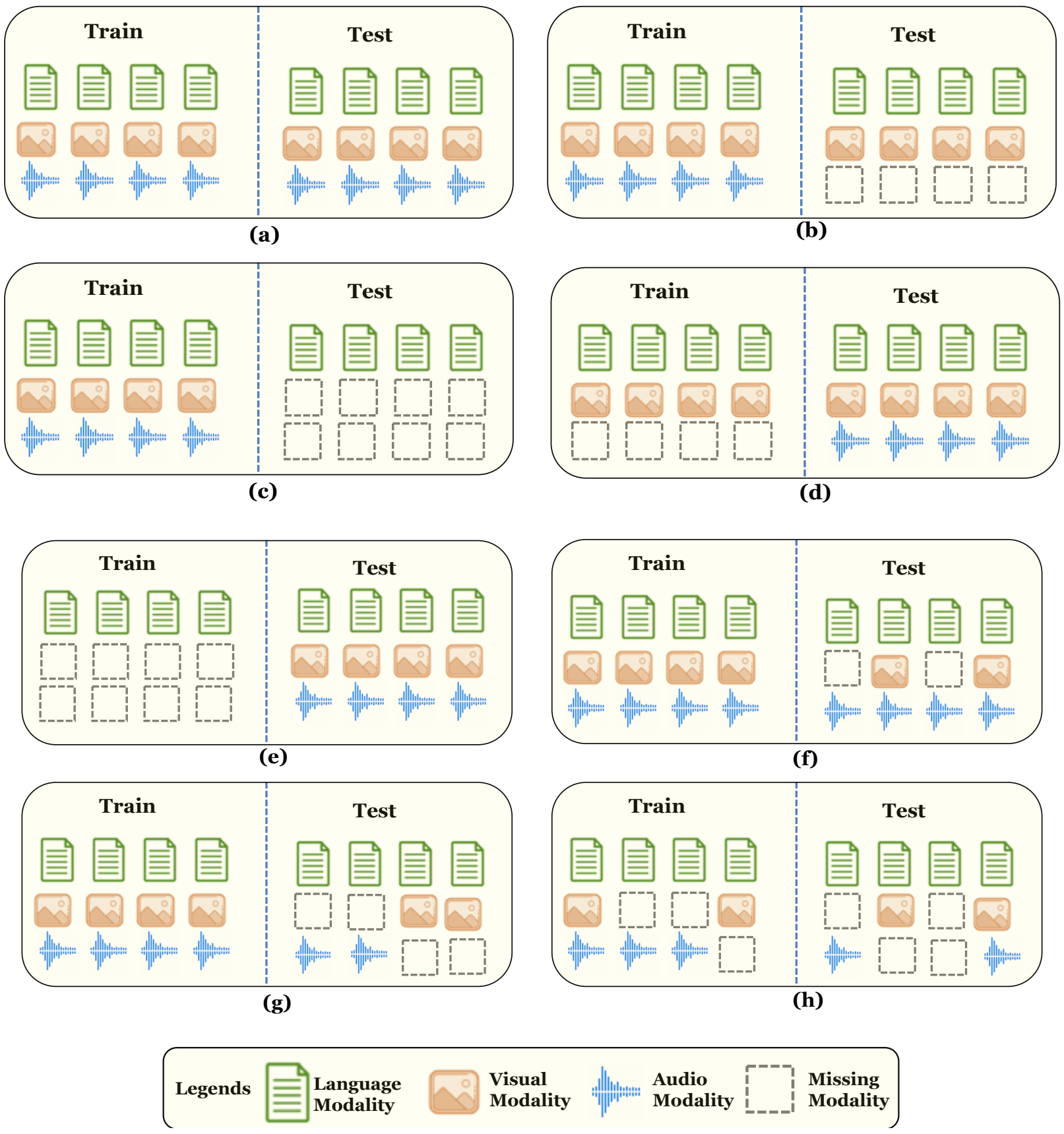

**Fig. 6.** Modality conditions at train and test time for three modalities - language, visual, and audio (enhanced based on [59])

If we consider a standard multimodal task consisting of multiple modalities, all the modalities are present at training and test time. In co-learning, all the modalities are present at training time and some are missing at test time. The modalities which are not present at the test time support other modalities during training. As a natural extension, there are multiple modalities during training time and only one modality is present during testing, effectively making it a unimodal scenario at test time. We also considered further sub-classification based on if the missing modalities are entirely missing or missing for a certain period or percentage. The scenarios in which one or more modalities are missing at training time but available at the test time are also possible. Those can be considered under the domain adaptation category of co-learning. In Fig. 6, we consider three modalities: text, video, and audio and presented this in detail.

The research studies also highlight that modality used at test time can be dominant or weak to understand if the weaker modality supports stronger or vice versa. Knowing the supportive modality leads us to think of another sub-classification - a) Stronger Enhancing Weaker (SEW) [21] and b) Weaker Enhancing Stronger (WES) based on which modalities are used at test time, as shown in Fig. 7. However, we have not considered this as an explicit sub-classification as this is applicable in all missing modalities scenarios unless the modalities have equal contribution or regulate each modality's contribution. For unimodal cases, missing data is normally addressed by missing at random (MAR) methods, which find a good relationship with available data to replace the missing samples.

However, in multimodality, the correlation between missing modality and available modality is complex and non-linear; even the relationship between missed entries and available entries for the same modality is complex. In the initial period of multimodal evolution, inferring missing modality from other modalities is achieved using probabilistic relations, with the models like Deep Boltzmann Machines [60]. Some techniques used modality imputations like cascaded residual autoencoder [61], imputing the kernel matrix of missing modality using kernel matrix of other modalities [62]. Lately, knowledge distillation like a teacher-student network [63] that learns from soft labels is utilized instead of imputation.

Deep learning-based approaches like autoencoders, adversarial learning, multiple kernel learning methods are used, and further enhancements are in progress. We included some of these recent approaches here.

#### *4.1.1 Fully missing modality*

The easier approach to implement missing modality at test time could be late fusion, which uses two unimodal networks and a weighing scheme to predict the results. If one modality is absent during test time, other unimodal networks can make a prediction. However, late fusion considers each modality as independent and cannot learn multimodal interactions [64]; hence it is not preferred for co-learning.

In some cases, co-training helped to handle missing modalities at test time. The co-training is extended to multimodal [47] to classify weakly labeled videos downloaded from the web with five modalities - RGB, motion, audio, concatenation of RGB-motion-audio, and metadata using decision level fusion. During testing, video metadata is not used as it is normally not available for video classification implementing missing modality at test time.

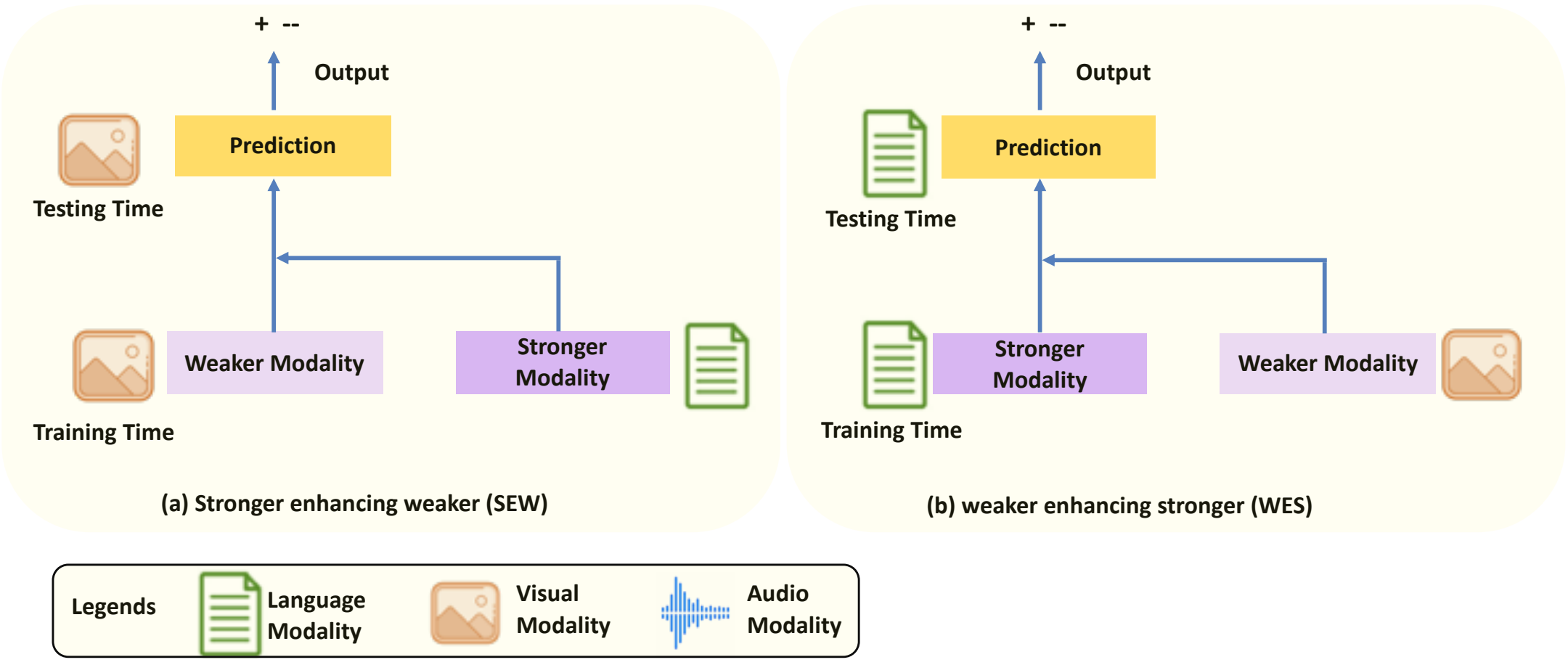

**Fig. 7.** Categorization based on dominant modalities at training time

Missing modality is a well-known problem in the domain where data from multiple sensors is processed. For example, remote sensing like all-weather mapping uses synthetic aperture radar (SAR) images and optical images; however, optical images may not be available due to poor weather conditions. Knowledge transfer and distillation approaches are typically used to handle missing modalities for multisensory data using image registration. Instead, sensor variant and invariant representations are learned to arrive at meta-sensor representation [9]. A prototype network is trained which, can generate a network for missing modality. Knowledge distillation uses a teacher-student network and privileged information learning [65] to have a multimodal distillation network for video action recognition using RGB and depth modality. A hallucination network is trained to mimic depth network using RGB input by training it in a teacher-student setup.

To reconstruct missing modalities, the researchers added cross-modality to autoencoder [24] with deep canonical correlation analysis (DCCA), referred to as DCC-CAE. Audio and visual modality is used in training, but only one modality is used at test time. Experiments on the CMU-MOSI [66] and CMU-MOSEI [5] datasets show better performance than the individual modalities as well as sequence-to-sequence multimodal models. Audio modality is a weaker modality, and both the combinations, i.e., audio at test time and video at test time, are evaluated.

The approach to factorizing multimodal data [67] into discriminative and generative factors reconstructs missing modalities. The discriminative factors are common across all the modalities and used for discriminative tasks, and generative factors are specific to modality and are used to generate modality data. This approach follows the reconstruction of missing modalities from the observed modalities and also increases the discriminative performance. The multimodal factorization model shows better performance when one modality is missing at test time than seq-2-seq models; language modality is seen as dominant and hard to reconstruct compared to audio and video modalities.

Vision-Language (VL) tasks like caption-image retrieval are trained on image and text modality; however, either image or text is provided as an input during test time, i.e., one modality is missing at test time. Also, cross-modal retrieval tasks in which the model is trained using image and text, but either text is used to retrieve image or image is used to retrieve text are examples of one full modality missing at test time. ViCo (word embedding from Visual Co-occurrences) using visual co-occurrences between object and attributes words [68] outperform text-only GloVe [69] embedding on VL tasks like image captioning, image retrieval, which have only text modality at test time. Performance is also evaluated on unsupervised clustering, supervised partitioning, and the zero-shot setting. Four types of co-occurrences are used between image and text – attribute-attribute, object-attribute, context, object-hypernym.

The sentiment analysis model is trained on visual, acoustic, and language modalities and tested only on language modality outperforms the individual language modality models [15]. This scenario is the case where *more than one modality is missing at test time*. Memory Fusion Network (MFN) consists of a system of LSTMs with one LSTM network for each modality, Delta Memory Attention Network for cross-modality attention, and Multi-view Gated Memory to store history of cross-modality interactions over time is used as an end to end model. Language modality is a dominant modality, and hence this can be classified as a WES sub-class as mentioned in Fig. 7. Co-learning implementation using MFN also proved domain adaptation by showing better performance on a different dataset but from the same domain and a different dataset from another domain.

Many multimodal sentiment analysis models employ language as a dominant modality while downplaying audio and video modalities. Hence, heterogeneous modality transfer learning (HMTL), which transfers knowledge of text modality as a source to audio-video modalities as a target, is proposed [70]. In this, text (source) modality representation is implemented to reconstruct audio-video modality (target) representations using a decoder to obtain a correlation between them. Higher sentiment classification performance is observed on CMU-MOSI [66] and IEMOCAP [71] datasets with HMTL. SoundNet [72] uses synchronization between audio and video in the unlabeled videos from the web instead of labeled data with a teacher-student network. A pre-trained ImageNet [45] and PlacesCNN [73] to extract objects and scenes from video frames is a teacher network, and the network for audio waveforms is a student network. This combination can transfer knowledge from the vision domain to sound without explicitly labeling videos for sound and used on sound modality at test time.

Another way to manage missing modalities at test time is to have required representations of modalities together. In the Multimodal Cyclic Translation Network model (MCTN) [23], a translation process from one modality to another creates a representation of both when trained with cycle consistency and prediction loss together. MCTN is extended hierarchically to have more than two modalities - audio, video, text. MCTN with only text modality at test time shows better performance than current state-of-the-art models including, co-learning implementation with MFN [15]. Like MFN, language is a dominant modality in MCTN and can be classified as WES.

Multitask model [6] is used to handle missing modalities. The model contains multiple classifiers, one for image, one for text, and another for prediction using the fusion of both modalities. This approach is mentioned as multi-task as it involves two monomodal and one multimodal classification for predicting the same tasks across three, i.e., prediction of sentiment. Since there are two separate classifiers – one for each modality, missing modalities at training and testing time is handled easily. This model structure enables the monomodal classifier to be trained with image-only or text-only modalities if one modality is missing at train time. They also used unpaired images to train the model, proving the model's ability to handle missing modality at training time. The model performs better when full multimodal data is used against the different proportions of multimodal data.

In another implementation with multi-task learning [17], an overlapping relationship between different types of VL tasks is utilized. A large-scale model is trained on 12 different datasets for tasks like visual question answering (VQA), caption-based image retrieval, grounding referring expressions, and multimodal verification. This discriminative VL model uses a different head for each task, like branches of a common, shared 'trunk' ViLBERT. The different tasks would overfit at different times; hence a dynamic stop-go mechanism is used based on validation loss. Image retrieval and image caption generation use only one modality at test time, thereby supporting co-learning objectives. The use of multiple datasets of different tasks is a step towards the generalization of a task.

Alignment and translation models together can be used for co-learning [21] to address the challenge of missing or poor quality modality at test time. A translator takes weaker modality features to generate stronger modality features; in the process, the encoder creates an intermediate representation, capturing information between modalities using a translation loss. The model performs better than unimodal on the RECOLA [74] dataset using audio modality at test time for valence and video modality at test time for arousal. Paired and annotated data is used for both modalities. The arrangement can be classified as a SEW sub-class.

Cross-Modal Cycle Generative Adversarial Network (CMCGAN) [75] consists of four encoder-decoder subnetworks, audio-to-visual, visual-to-audio, audio-to-audio, video-to-video creating four generations paths including four discriminators. This arrangement helps to generate any of the absent modalities. The use of shared latent space of modalities and a generator for each modality to create relationships between modalities is proposed [11] with adversarial learning for generating missing modalities on ImageNet-EEG [76] dataset.

Thus, these representative implementations highlight the multiple methods to handle the fully missing modality scenarios. These implementations cannot be compared effectively as each method has pros and cons and is created for different tasks and datasets.

#### *4.1.2 Partly missing modality*

In the above section, we discussed the scenario in which one or more modalities are fully missing at test time. However, consider an audio-visual application with the following two scenarios - a) the video camera is not working for a brief span, or there is a darkness at the user place, but the audio is clear. Here, we cannot capture facial expressions needed for detecting emotions as visual modality is missing. Vice-versa, b) the camera is available, but the audio is not available for a short span. Here, we cannot consider audio to measure sound variations. The percentage of missing data for a modality could vary from mildly missing to severely missing. The co-learning model should be designed to work in these situations of *partly missing modality* with optimal performance.

A multi-style training method is used in the speech enhancement multimodal model [77]. The audio-visual, visual-only, and audio-only inputs are selected randomly for the defined number of epochs during training. This arrangement assists in studying partly missing modalities for varying duration at test time and identifying the dominant modality and its contribution to overall performance. Multimodal learning with severely missing modality (SMIL) [59] approach considers severely missing scenarios (i.e., 90% modality is missing). It shows if a model can support the missing modalities in training, testing, or both, and the model still produces comparable results. The Bayesian meta-learning framework is proposed by having two auxiliary networks, one for reconstruction and another for regularization. The experiments on Multimodal IMDB [78], MOSI, and Audiovision MNIST [25] outperform other methods like variational autoencoder (VAE) and GAN when one modality is fully available, and another modality is severely missing. Other combinations such as a) training with full image + n% of audio and testing with only image and b) training with full image + n% of audio and testing with full image + full audio have been experimented.

Thus, a partly missing modality at training or testing time poses a different challenge than a fully missing modality in which we consider the fully missing modality as the supporting modality. On the other hand, in the case of partly missing modality, we need to design models to support each other with a bi-directional transfer of knowledge warranting different training mechanisms.

*4.1.3 Discussion*

The presence of modality is a key consideration for multimodal co-learning. This situation is common in the real world and needs to be addressed for the robustness of multimodal models. With the progress in VL tasks and media retrieval tasks, the scenario of only one modality at test time is prevalent now. However, it is not explicitly mentioned as co-learning, and it is considered part of the alignment, translation, and representation learning. We defined sub-classes for the presence of modality based on full or part presence of it. We also highlighted that there are instances where more than one modality is missing. The modalities are missed at a testing time as well as training time partly or fully. There is generally one dominant modality that helps us to decide supporting modalities. The details explained in this section are summarized as highlights in Table 1.

It is evident that the ability to handle missing modalities depends on the amount of co-learning achieved among the modalities. Hence, unimodal methods extended to handle missing data in multimodal applications are not sufficient. Autoencoder-based reconstruction and multimodal data factorization utilized co-learning among modalities and performed better than the extension of unimodal methods for multimodal applications. Attention models, multimodal embedding, transfer learning, and cyclic translations further enhanced co-learning among the modalities. Handling missing modalities at training time is challenging; however, GAN and meta-learning based frameworks seem to be a step forward.

**Table 1**: Summary of studies for the presence of modality co-learning objective

| Presence of modality | Sub-class | Multimodal Taxonomy Area | Summary of the studies | | | | Observations and features | References |
|---|---|---|---|---|---|---|---|---|
| | | | Modalities at Train | Modalities at Test | Applications | Methods | | |
| Fully Missing | | Representation, Co-learning | Image, Text | Image | Image Classification | Multiple Kernel Learning | Semi-supervised and weakly supervised and needs multi-stage training. | [48] |
| | | Representation, Co-learning | Audio, Video, Text | Audio, Video | Video Classification | Co-training | Weakly supervised, classifier for each modality & a voting mechanism. | [47] |
| | | Representation, Co-learning | Audio, Video | Audio | Sound Classification | Teacher Student | Weak supervision using synchronization between audio and video. Image classification with ImageNet, PlacesCCN, and mapping with associated sound. | [72] |
| | One - Full missing | Representation, Co-learning | SAR* Images, Optical Images | Optical Images | Remote Sensing | Prototype Network | Meta-sensor representation with knowledge distillation. Use for audio, video, and text modality will be challenging. | [9] |
| | | Representation, Co-learning | RGB, Depth | RGB | Video activity Recognition | Knowledge Distillation and Privileged Learning | Supports noisy conditions, needs hallucination network per modality. | [65] |
| | | Representation, Fusion, Co-learning | Audio, Video | a) Audio b) Video | Sentiment Analysis | DCC - CAE (autoencoder) | Reconstruct of missing modality, and CCA is used for coordinated representation. | [24] |
| | | Representation, Fusion, Co-learning | Audio, Video, Text | Three pairs: audio-video, audio-text, video-text | Sentiment Analysis | Multimodal Factorization | Discriminative (common) and generative (specific) factors are obtained for modality reconstruction. | [67] |
| | | Representation, Alignment Co-learning | Image, Text | Three tasks: image, text, image-text | Vision-Language Tasks | Multimodal Embedding | Visual co-occurrences between objects and attributes words. | [68] |
| | | Representation, Alignment, Translation, Co-learning | Audio, Video | a) Video b) Audio | Sentiment Analysis | Encoder-Decoder with alignment | Translation of weak modality to stronger creates representation to be aligned with a representation of stronger ones using CCA. | [21] |
| | | Representation, Co-learning | Image, Text | Three tasks: image, text, image-text | Vision-Language Tasks | Multi-task Learning | Multi-task using ViLBERT using 12 datasets and dynamic stop-go schedule for training. | [17] |
| | | Representation, Fusion Co-learning | Image, Text | a) Text b) Image | Emotion Prediction | Multi-task Learning | Multi-task model can handle missing modality at test and training time. Later enables the use of un-paired images in training. | [6]# |
| | | Representation, Alignment, Fusion | Audio, Video, Text | Audio, Video | Sentiment Analysis | Transfer Learning & Adversarial Learning | Text is used to reconstruct audio-video modality | [70] |
| | | Representation, Alignment, Fusion | Audio, Image | Audio or Image | Image sound generation | Cycle GAN | Four encoder-decoders in a generator-discriminator arrangement | [75] |
| | | Representation, Fusion | EEG, Image | EEG | Cross-modal Retrieval | Triangle GAN and Cycle GAN | Shared latent space is divided into semantic and semantic-free latent variables with | [11]# |
| | | | | | | | modality specific generators. | |
| | More than one modalities are missing | Representation, Fusion, Co-learning | Audio, Video, Text | Text | Sentiment Analysis | Memory Fusion Network | Supports domain adaptation, all aligned data, no modality reconstruction is required. | [15] |
| | | Representation, Alignment, Translation, Co-learning | Audio, Video, Text | Text | Sentiment Analysis | Translation with Encoder-Decoder | Translation between modalities with cyclic translation and prediction loss. Supports noisy modality. | [23] |
| Partly | Alternately missing | Representation, Fusion, Co-learning | Audio-Video alternate | Audio | Speech Enhancement | Encoder-Decoder using CNN | Supports noisy data, needs synchronization between audio and lip movement. | [77]# |
| | | Representation, Co-learning | a) Image, Text b) Audio, Image | a) Text (0-90% missing) b) image | a) Sentiment Analysis b) audio-visual classification | Meta-learning framework | Severely (~90%) missing modality case is evaluated using a modality-complete dataset as prior. The performance is at par with VAE & GAN. | [59]# |

*SAR - synthetic aperture radar images
# Note: In all the above studies, modalities are missing at the testing time; in [39], [58], [62] and [64] modalities are missing at training and testing both.

### *4.2 Data parallelism*

The modality data where a direct or strong alignment among modalities is observed is referred to as *parallel data*, e.g., in a speech dataset where each text and audio sample is aligned at a word level. In *non-parallel* data, there is no direct link between modalities at an observation level, but there is an overlap in terms of category. For example, audio-video-textual samples in an instructional video may not be directly linked to each word. Still, they may be related at the step level, video level, or video segment level [79]. When the two modalities are associated through a shared modality or a dataset, it is termed as a *hybrid data* approach, e.g., in multilingual image captioning; image modality would be paired with one language. Since data parallelism is dependent on a pairing of data between modalities, it can also be termed as a strongly paired and weakly paired data approach. Three types of data parallelism are depicted in Fig. 8.

#### *4.2.1 Parallel data or strongly paired modalities*

In this, all modalities samples are aligned among themselves, e.g., images and captions are paired in the image-captioning multimodal application, as shown in Fig. 8a. In another example, audio segments, video frames, language words are paired in multimodal sentiment analysis. The various VL tasks such as image captioning, scene description, video captioning, media retrieval, etc. [17] use paired modalities data at training time and only one modality test time, supporting the co-learning principle.

Translation-based co-learning methods also demand strongly paired or parallel data [21] to achieve sentiment classification. The speech enhancement multimodal model [77] uses visual modality as a supporting modality to enhance audio modality using either lip sync visual data or facial expression data, strongly paired with audio utterances.

The use of parallel data or strongly paired data is an intuitive approach for one modality to support other modalities with the transfer of knowledge. The multiple applications use parallel data for training and testing for multimodal tasks. Researchers are working to find alternatives as creating strong pairs is costly and time-consuming.

*4.2.2 Non-parallel data or weakly paired modalities*

Preparing parallel data is not efficient, and efforts increase with an increase in the number of modalities. Moreover, creating strong pairs needs offline pre-processing, thereby restricting the model's end-to-end learning capability. There are chances that data is not aligned across modalities at finer levels but may be aligned at a coarse level. The first path-breaking approach started with the use of self-attention models that got popular in language modeling to handle long sequences. Promptly, attention is extended to cross-modality, creating automatic alignment between modalities, e.g., Co-attention maps words and image regions for phrase grounding creating alignment between different modalities for VQA tasks [80]. *Hierarchical co-attention* [81] is designed for attention at word, phrase, and a sentence level between question and image and then recursively combined from word to sentence level to obtain higher performance in the VQA task.

Emotion recognition on Automatic Speech Recognition (ASR) is performed [82] by using an attention mechanism to align the transcript of text and audio signals than manually pairing data. The multimodal co-learning objectives have not been experimented with in the above attention-based implementation and would be of interest to check how the models perform for missing modality.

Non-parallel data or weakly paired data approach does not require paired data; it utilizes a shared concept between the modalities, as shown in Fig. 8b. The *deep visual semantic embedding* model (DeViSE) [83] uses text to get better visual representations by coordinating text embedding features with convolutional neural network (CNN) based image features. At test time, the nearest neighbor of visual representation is found by referring to the embedding space for a new image. The model uses some paired data, and the rest is non-paired data to arrive at semantic embedding.

End-to-end training on uncurated instructional videos using Multi-Instance Learning with Noise Contrastive Estimation (MIL-NCE) loss obtained joint embedding of video and text on HowTo100M dataset [79]. There is no strong pairing between video and text; however, both the modalities are aligned at the instructional video level. Action recognition, action-step localization, action localization, and action segmentation tasks are evaluated on text-to-video retrievals, which has text query at test time, thereby fulfilling the objective of co-learning.

The recent advent of techniques that can process long temporal sequences [51] and store context for a longer duration enables the use of non-parallel data for multimodal co-learning, providing a level of performance as that of strongly paired data.

*4.2.3 Hybrid data or bridged data modalities*

A shared modality or a dataset is used to create a pairing between two modalities in a hybrid data approach. Hybrid data can also be called a bridge, as often a bridge is created between two modalities with the help of a dataset or a modality. Fig. 8c shows a one-to-one bridge between modalities with the image as a pivot modality. Multiple datasets are available with images and captions in English. In contrast, the same number of datasets are not available with images and their captions in Russian, Hindi, Urdu, etc. These scenarios are addressed by considering the image as a pivot view. One can create translation from one language to another using networks like Bridge Correlational Neural

Networks (Bridge CorrNets) [10]. The network learns aligned representations across multiple views using a pivot view and can be generalized to multiple modalities with one pivot modality, as shown in Fig. 8d.

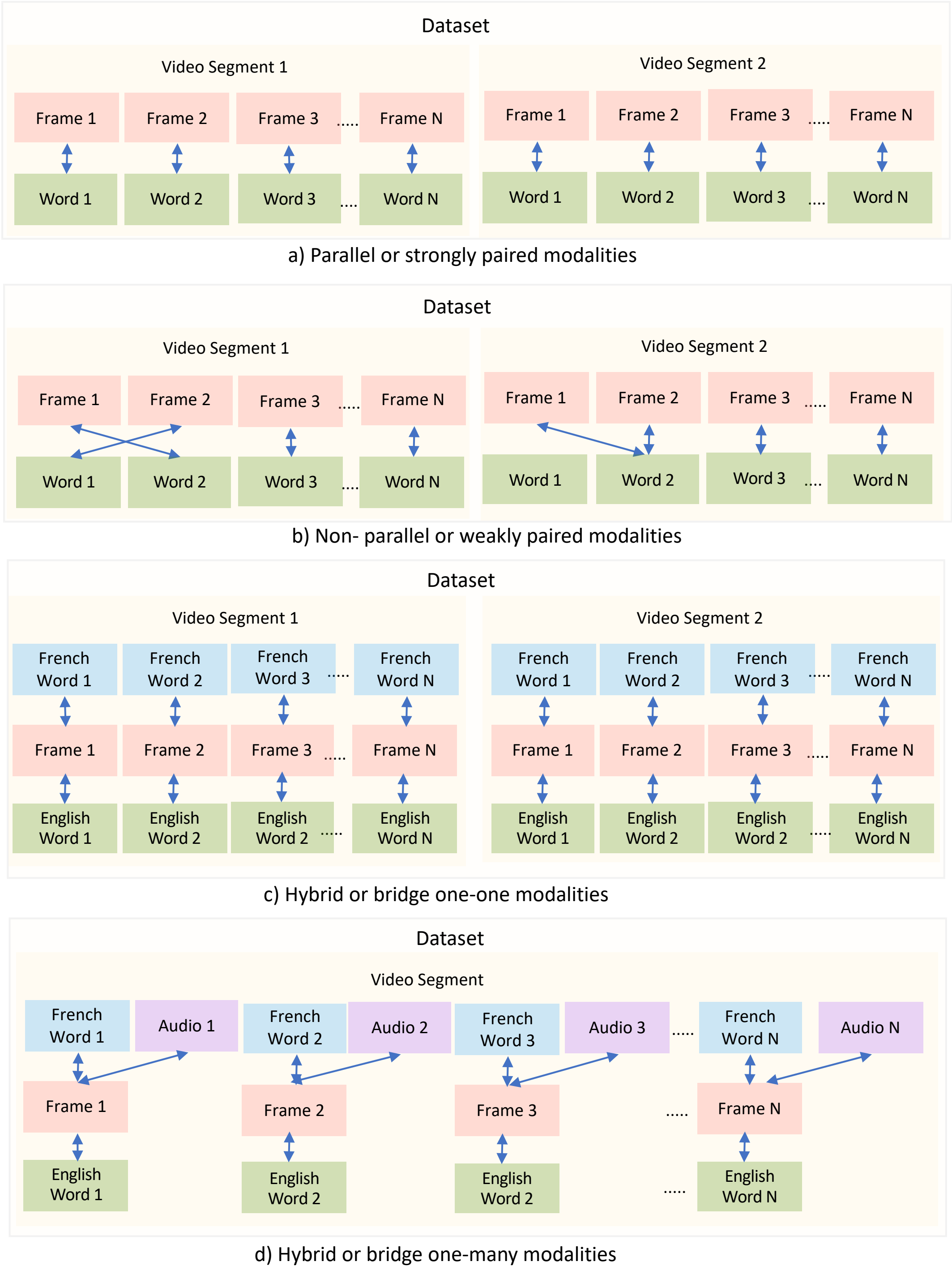

**Fig. 8.** Classification of data parallelism in multimodal co-learning (enhanced based on [13])

Bridge relations are used to build multimodal and cross-lingual embeddings [84] for video search using multilingual text. Video modality is considered as a pivot modality, which got an inherent relationship with multilingual captions. This relationship can serve as an alignment and supervised data to create multimodal multilingual embedding; contrastive learning methods help form intra-modal, inter-modal, and cross-lingual pairs among data points. Transformers trained using video and multilingual text has outperformed transformers trained on just video and English text for video search using multilingual text. They performed better at zero-shot learning.

In the year 2016, the conference on machine translation (WMT16) [85] had created a challenge for two tasks- a) create a translation of image description to the target language with or without clues from an image, b) generate the description of an image in the target language with or without clues from source language description, i.e., in first task image can be a pivot, and in the latter task, source language can be a pivot to have an aligned data. Participants proposed multiple methods like Recurrent Neural Network (RNN) and Region-based Convolutional Neural Networks (R-CNN), encoder-decoder with attention, and pre-training of models on another dataset. Although multimodal results were not encouraging, this set a direction for using shared relationships among source and target. This approach is classified as hybrid one-to-one, as shown in Fig. 8c.

There is an inherent synchronization between audio and video in a video clip that can be utilized as bridge alignment. SoundNet [72] uses this relationship by extracting objects and scenes from video using pre-trained image networks, which act as teacher network and audio data as an input to student network. In this teacher-student arrangement, knowledge transfer from video to audio becomes possible, thereby obtaining sound representations. Here audio representations of 2 million sound clips are obtained using unlabeled videos.

Instructional videos and textual instructions are readily available over the web for cooking recipes. The sequence of instructional steps in the two recipes is different among the text and video recipes. But, the relations between videos and textual instructions of the same recipe can be aligned by mapping to pivot modality. Video transcript and textual instructions can be aligned to arrive at the mapping between a video and textual instructions, which are different recipes of the same dish [86]. This approach is used to create a large Microsoft Research Multimodal Aligned Recipe Corpus dataset containing 150,000 pairwise alignments between recipes across 4,262 dishes with rich commonsense information. In other instances, objects in a video are used to create the mapping between source and target videos, thereby aligning instructions. This approach can be classified as one-to-many hybrid data parallelism, as shown in Fig. 8d.

Multiple studies show that multimodal models can have better word representations, i.e., embeddings, than just text embeddings. One way to create a multimodal embedding is to have a projection of aligned multiple modalities data into a common sub-space governed by a similarity matrix. In the absence of aligned modality data, alignment is created using shared pivot information among the modalities. Associate Multichannel Autoencoder [87] was used to learn associations among modalities using reconstruction of modalities and use associations of words to create a mapping between conceptual information, i.e., audio, video. Three different datasets, Glove for text, updated ImageNet with WordNet [88] synsets, Freesound for audio, and a word association dataset, are used for training. The model is tested on word similarity and word relatedness and showed a better performance than textual embeddings and other multimodal embeddings. This approach can be classified as one-to-many hybrid data parallelism, as shown in Fig. 8d.

Thus, hybrid data can take advantage of large-scale data from the web with pivots or bridges generated through people's actions. For example, captions for a movie in multiple languages create a bridge between various languages with video as a pivot modality.

*4.2.4 Discussion*

Multimodal co-learning methods should support three types of data parallelism for real-life conditions. The details explained in this section are summarized in Table 2 for data parallelism. Data parallelism can be further sub-classified based on data conditions at training and testing time as there could be differences. It is observed that with the latest techniques, non-parallel data, as well as hybrid data, are used for multimodal co-learning instead of always looking for strongly paired data which is costly and time-consuming.

With the advent of internet technologies, huge data with weak relations or shared relations is getting generated and can be used. Relationship among modalities data points needs to be established in case of weakly paired data. Attention-based methods identify specific information which can be held in the temporal sequence. Attention types such as cross-modality attention, co-attention, and hierarchical attention proved helpful to create relationships among the modalities. Multimodal embedding, which creates associations among similar data points, is also used. The known shared relationship among the modalities, for example, language translations, objects in videos, steps of recipes, help to create a pairing of data points required to achieve co-learning.

**Table 2**: Summary of studies for data parallelism objective of co-learning

| Data Parallelism | Sub-class | Multimodal Taxonomy area | Summary of the studies | | | Observations and features | References |
|---|---|---|---|---|---|---|---|
| | | | Modalities | Applications | Methods | | |
| Parallel or strongly paired | NA | Representation, Fusion, Co-learning | Audio, Video, Text | Sentiment Analysis | Memory Fusion Network | All three modalities are aligned (strongly paired) at word level by P2FA[1] . | [15] |
| | | Representation, Co-learning | Image, Text | Image-Captioning, Scene Description, Media retrieval | Multi-task model with ViLBERT | Data from all 12 datasets are aligned (strongly paired). | [17] |
| Non-parallel or weakly paired | NA | Representation, Co-learning | Image, Text | Image Classification | Multimodal Embedding | The deep visual semantic model uses non-parallel data using the semantic relationship between image labels and word embeddings. | [83] |
| | | Representation, Co-learning | Video, Audio | Sound Classification | Teacher-Student | Synchronization between audio and video is used as a bridge. Image classification with ImageNet, PlacesCCN, and mapping with sound on large unlabeled video data. | [72] |
| | | Representation, Co-learning | Video, Text | Text-Video retrievals | Multi-Instance Learning with contrastive loss | Data is aligned at video and instruction-level and not at word or frame | [79] |
| Hybrid (Bridge) | One - One | Representation, | Image, Text | Language Translation | Bridge Correlational | Pivot modality is image, source and | [10] |
| | | Alignment, Co-learning | | | Neural Networks | target modalities are text only. | |
| | One - Many | Representation, Alignment | Video, Text | Alignment Task | Component Level Alignment (object in videos – text in the recipe) | Multiple videos and textual recipes of a dish, thereby creating one-many relation. Alignment is created between text-text, text-video, and video-video recipe pairs. | [86] |
| | | Representation, Translation, Alignment | Video, Text (9 languages) | Multilingual Video Search | Multilingual Multimodal Transformer | Video as pivot modality for nine languages and contrastive learning used for multimodal embedding. | [84] |
| | | Representation, Co-learning | Image, Text | Multilingual Machine Translation | Encoder-Decoder with attention | Image is a pivot modality for translation, and the source language is a pivot for cross-lingual image description. | [85] |
| | | Representation, Fusion, Co-learning | Audio, Video, Text | Multimodal Word Representation | Multimodal Embedding | Association of words is used to create a mapping with audio-video. Different weight to each modality for a word | [87] |

[1] https://github.com/srubin/p2fa-vislab

### *4.3 Noisy Modality*

Another challenge addressed by multimodal co-learning is robustness and performance in the presence of noisy conditions. Noise could be of two types- a) Label Noise b) Data Sample Noise. Earlier, techniques like teacher-student, co-sampling, co-teaching, as mentioned in Section 3, were used to handle label noise and data noise. Recently deep learning-based techniques such as adversarial networks, generative networks, etc., are showing promising results.

#### *4.3.1 Label noise*

*Label Noise* can be due to human errors by annotators, lack of expert annotators, the complexity of annotations (e.g., medical imaging), faulty annotating conditions, poor quality of data samples, subjective classification, use of associated meta-data for classification, or even cost concerns due to which weak annotations are sought. The noisy label conditions are normal in real life, contrary to the assumptions during model development. Mislabeled data affects the application performance, needs more training data, and may create class imbalance. Label noise is classified into three types [89], as described in Table 3. Other label categorizations can be based on the level of noise. There are also some occasions of out-of-distribution label noise when an incorrect label corresponds to an unknown class that is not in a dataset.

**Table 3**: Classification of label noise [89]

| Abbreviation | Label Noise Category | Description |
|---|---|---|
| NCAR | Noisy Completely at Random | Noise is random and does not depend on instance features or true class labels. It is also referred to as symmetric noise as it is the same for all classes. |
| NAR | Noisy at Random | Noise is independent of the label's feature but depends on its class, e.g., certain classes may have more incorrect labels. |
| NNAR | Noisy Not at Random | Noise depends on the feature and a true class label. |

The simplest approach is to remove samples that appear mislabeled [90]; other methods include looking at the impact of those labels on classification performance and remove them. A semi-supervised approach is used by removing only labels but keeping samples and then reclassifying them. However, detecting label noise is difficult; instead, we can have machine learning models to handle label noise. These models should look at classification accuracy, model complexity, and estimation of noisy labels. Deep learning models are error-prone as they learn along with noise conditions quickly and overfit on it. Label noise needs to be handled during training time so that trained models can gain the required performance level.

Domain knowledge helps to address label noise by understanding inherent structure among different data classes. An ontology-based approach uses a hierarchical structure among different sound categories for classification tasks. Multi-task Graph Convolution Network (MT-GCN) [91] model is trained on well labeled and noisy-labeled samples and graph network creates an ontology between the two tasks encoding label relationship. Conditional GAN (cGAN) [92] and auxiliary classifier GAN (AC-GAN) [93] are the extensions of GAN, which selectively generates data based on class labels. Label noise-robust GAN (rGAN) [94] adds noise transition module in which discriminator finds decision boundary between real and generated data on noisy labels, and generator tries to generate data non-distinguishable by the discriminator.

Adding symmetric noise to labels of test samples for few-shot learning, Cross-Modal Alignment (CROMA) [95] based modality generalization model shows more robustness. The following datasets with label noise are published to investigate label noise conditions – a) Clothing1M [96] dataset contains 1M images for 14 classes of clothes with noisy labels and 74,000 clean labels across train, validation, and test split, b) WebVision [97] dataset contains 4.5M million images from Flickr [98] and Google with tags, captions, etc., across 1000 classes, and 50 images per class as clean labels.

Thus, methods to handle label noise evolve from earlier co-teaching methods to the latest GAN-based methods, making models ready to manage label noise at inference time to meet the co-learning objective.

#### *4.3.2 Data noise*

As discussed in Section 3, student-teacher networks increase robustness against noisy conditions in unimodal and multi-view learning models. Multimodal deep learning models are robust than unimodal models as they use supplementary information among the modalities. For image modality, noise can be classified into two types - *a) adversarial perturbations b) common corruptions* [99], which can be extended to other modalities. Common corruptions include Gaussian noise, real-life variations like snow, rain, smoke, motion, zoom blurs, etc. For further study, corrupted versions of standard datasets [100] - ImageNet-C, Tiny ImageNet-C, and CIFAR10-C are published and used. The physiological signals [101] from wearable devices, including electroencephalogram (EEG) and electrocardiography (ECG), are usually noisy, susceptible to environmental interferences, non-stationary, and require time series analysis. This section covers how multimodal models provide robustness against data noise using co-learning and how the techniques being used here support missing modality and other objectives of co-learning.

Speech enhancement (SE) tasks focus on improving speech quality by reducing noise from the noisy speech input. Audio-visual multimodal models for speech processing showed higher performance over only audio modality. Convolutional Neural Network (CNN) based model is used [77] to process audio and visual modality data separately and then fused. Noisy speech and visual data are used for training as input, and clean speech and visual data are used at output, forming an encoder-decoder structure of a model. This structure created a multi-task model with reconstructed visual output as auxiliary output supporting missing visual modality at test time. Ambient and interference noise added with different noise to interference ratios. Multiple types of noise are used during training and testing to test the model's robustness to unseen noise types. A multi-style training method is used to study the impact of the dominant modality. The audio-visual, visual-only, and audio-only inputs are selected randomly after a defined number of epochs. Here, audio modality noise is addressed using visual modality as a supporting modality to audio modality in a co-learning arrangement.

With adversarial examples on a convolution attention network for audio-visual event recognition [102], the robustness of fusion techniques early/late/hybrid is presented. The perturbations are added only to the audio modality of the Google Audio Set dataset. Adversarial noise at low frequencies tends to have higher attack potential than at higher frequencies, even though low frequencies features are not stable. Classes trained on more data have the robustness to attack than classes trained on fewer data for an imbalanced dataset. The late fusion proved to be robust compared to early and hybrid fusion. A trade-off between robustness and accuracy is crucial while choosing the architecture of a neural network.

Pre-trained VL models are more robust than their task-specific state-of-the-art models. But this robustness is still limited and demands techniques to increase the same. Multimodal Adversarial Noise GeneratOr (MANGO), which adds the noise in the embedding space of VL models, is proposed [103] to study four types of the robustness of VL models, namely, robustness against a) linguistic variation; b) logical reasoning; c) visual content manipulation; and d) answer distribution shift using nine diverse datasets. Rather than creating local perturbations that eventually get learned by the model, MANGO is a neural network-based noise generator that won't allow a model to adapt to it. Additionally, some regions of input images are masked, and some tokens of text are dropped to increase diversity as it impacts distribution. Models trained using MANGO show better performance on current benchmarks.

There is good progress in handling noisy data, particularly in the visual modality domain, including adversarial noise generation for images. These matured techniques are now studied for multimodal data, and adversarial techniques are the leading ones.

#### *4.3.3 Data imbalance*

One of the challenges of machine learning is the lack of balanced data across all the classes, and the same is applicable for multimodal models. *Class imbalance* leads to reduced model performance, over-representation of majority classes, and sometimes minor classes are ignored or treated as noise by classification models. Some of the earliest techniques to handle data imbalance are oversampling, under-sampling, the synthetic minority oversampling technique (SMOTE) [104], and adaptive synthetic sampling (ADASYN) [105]. SMOTE and ADASYN both generate synthetic samples based on a linear combination of two nearby samples instead of following the real distribution of a minority class, and noise gets added when the boundary between major and minor classes is unclear.

In GAN, the generator learns the latent distribution of real data and produces real-like fake samples. Deep Multimodal Fusion Generative Adversarial Network (DMGAN) [106] uses a GAN to generate samples for each modality and a discriminator for each modality to discriminate real from fake samples. Overall fusion with inputs from all real samples and generated fake samples is used to maintain joint distribution. This model is used to classify faculty web pages using multimodal features – text, image, and HTML layout. The model can extend to missing and noisy modalities.

Although data imbalance is always viewed from a class imbalance perspective, there could be data imbalance among the modalities. GAN framework is used to generate visual features from the textual descriptions to balance the data in both modalities [107] to classify images using only visual data at test time.

#### *4.3.4 Discussion*

Initially, noise handling and creation of noisy data focused on images, i.e., visual domain; now, this is expanded to other modalities like text, audio, EEG, and physiological data. Each modality has different noisy conditions based on modality feature space, and there are various ways to generate the noisy data. Noise generators with adversarial learning generate noise based on local image distribution and global and universal features across the modalities or dataset. The adversarial techniques also overcome the challenge of deep learning models learning noise as features. Common corruptions like a blur, low light conditions, background noise get added when a real-life scenario dataset is prepared. The summary of representative studies is presented in Table 4. Similar to unimodal ones, noisy datasets are required for multimodal applications also. Co-learning among the modalities increases the robustness of multimodal applications compared to unimodal applications. GAN-based methods are the leading ones to handle label noise, data noise, and data imbalance. More studies are required to evaluate robustness to missing, noisy, and imbalanced modalities together.

**Table 4**: Summary of studies for noisy modalities co-learning objective

| Noisy Modality | Sub-class | Multimodal Taxonomy Area | Summary of the studies | | | Observations and features | References |
|---|---|---|---|---|---|---|---|
| | | | Modalities | Applications | Methods | | |
| Label Noise | Random/ Symmetric | Representation, Fusion | Audio, Ontology | Audio Tagging | Multi-task Graph Convolution Network | An ontology-based knowledge graph using the hierarchical structure of sound categories. | [91] |
| | | Representation | Image | Image Classification | Label noise robust GAN (rGAN) | cGAN, AC-GAN use label conditions. The noise transition module uses adversarial learning. | [94] |
| | | Representation, Alignment, Co-learning | Image, Audio | Audio Classification | Cross-Modal Alignment – Meta-Learning | Robust for 0-60% symmetric noisy labels for a few shot learning and domain generalization using meta-learning. | [95] |
| Data Source Noise | Adversarial noise | Representation | Image, Text | Vision-language Tasks | Adversarial Learning with Transformers | Noise Generator adds noise in embedding space of VL models with image masking and text token dropping. | [103] |
| | | Representation, Fusion, Co-learning | Audio, Video | Audio Visual Event recognition | Convolutional Attention network | Adversarial noise is added to the audio. Robustness-accuracy trade-off evaluated for fusion models. | [102] |
| | Common Corruptions | Representation, Fusion, Co-learning | Audio, Video | Speech Enhancement | Encoder-Decoder using CNN | Ambient and interference noise as input and clean speech visual data as output in encoder-decoder. | [77] |
| Data Imbalance | NA | Representation, Fusion | Text, Image, HTML layout | Web page classification | GAN | GAN generates samples for imbalanced data, and fusion preserves the joint distribution. | [106] |
| | NA | Representation, Co-learning | Text, Image | Image classification | GAN | GAN generates samples for imbalanced data of visual modality | [107] |

### *4.4 Modality annotations*

Preparing the required amount of labeled data is challenging and costly. Multimodality adds further difficulty as each modality may need separate annotation and an expert. Hence, unsupervised, semi-supervised, and weakly supervised learning techniques evolved. These are well established for individual modality data and are being explored for multimodal data.

#### *4.4.1 Supervised learning*

In supervised models, labels are available for all the samples in training data for all modalities. The initial multimodal implementations started with supervised learning using labeled data. Audio segments, video frames, and words are paired and labeled for multimodal sentiment analysis. Memory Fusion network (MFN) is trained using labeled data of three paired modalities while tested only on language [15].

The various VL tasks such as image captioning, scene description, video captioning, media retrieval, etc. [17] use paired and fully labeled data in training and only one modality at test time, supporting the co-learning principle. Labels are created in the form of captions, descriptions, or class names of images. Translation-based approaches for co-learning need labeled and strongly paired data [21] to achieve sentiment classification. The speech enhancement multimodal model [77] uses visual modality as a supporting modality to enhance audio modality which uses fully labeled utterance data for arousal and valence. Thus, supervised learning-based multimodal models are able to achieve co-learning objectives, as shown through studies in this section and previous sections.

#### *4.4.2 Semi-supervised learning*

In semi-supervised models, a small volume of data is annotated along with large un-annotated data. However, for multimodal models, two scenarios arise- a) some amount of annotated data is available in all the modalities, b) One modality data is annotated, and another modality data is un-annotated and vice versa. There can be multiple combinations based on the number of modalities being used. The earliest approach for semi-supervised learning is proxy label methods, which generate proxy labels on unlabeled data using a prediction function trained on the small labeled dataset. These methods are divided into two main classes self-training and multi-view learning, as discussed in Section 3.

An approach similar to co-training is adopted to classify images using images and their tags in a semi-supervised manner with a small set of labeled and rest unlabeled image-tag pairs [48]. Images and tags are treated as two modalities, and tags are used as a supporting modality to classify the images using a multiple kernel learning classifier. Deep Visual Semantic Embedding (DeViSE) [83] consists of labeled images and uses semantic information from large unannotated text data. It creates a semantic relationship between labels and textual data. Here part of the data is annotated, and another part is unannotated. When an image is presented for predicting labels at test time, its visual representation is used to look for the nearest label in the embedding space. The model also supports ZSL for the labels never observed visually.

#### *4.4.3 Weakly-supervised learning*

In weakly supervised models, annotations at a higher level are used to perform a task on the low level of data without explicitly creating labels at the lower level. Phrase grounding establishes a mapping between image parts and language phrases. Multimodal Alignment Framework (MAF) [22] uses image-caption mapping to arrive at the mapping between objects in the image and phrases using contrastive learning. A pre-trained object detector is used to extract objects from an image and predict labels, attributes, and features to form visual features. Textual features are obtained from captions, and an attention mechanism is used to have visually aware language representations. Contrastive loss maximizes the similarity scores between paired image-caption samples and minimizes the score between other negative samples. This study shows that the weakly supervised techniques for multimodal tasks can be extended to co-learning objectives. Likewise, to achieve contrastive learning, negative samples are created by substituting words in captions [108] and then maximize mapping between regions and corresponding captions compared to non-related areas and captions for a weakly supervised phrase grounding problem.

High-level relationships are used as weak supervision in multimodal co-training [47] to classify weakly labeled videos downloaded from the web using five modalities - RGB, motion, audio, concatenation of RGB-motion-audio, and metadata. Video metadata is used as weak supervision between video content and video class.

The multimodal content on the web is increasing with social media and content sharing platforms, and there is inherent weak supervision between multimodal content. Hence, co-learning with weak supervision is an important area for the future.

#### *4.4.4 Un-supervised learning*

In un-supervised models, training data is not labeled, and there are multiple approaches to use un-annotated multimodal data for various tasks. MAF for phrase localization is one way to achieve an un-supervised multimodal model [22]. In MAF, a pre-trained object detector is used to extract objects from images, and their labels are mapped with textual captions using contrastive objectives. Obtaining real-world labeled high stake deception detection data is difficult and costly. Hence, unsupervised audio-visual subspace transfer learning is proposed [109] to detect high-stakes deception using easily available lab-controlled low-stake deception data. This method is categorized as Unsupervised Domain Adaptation (UDA), which uses labeled source domain knowledge to perform tasks in the unlabeled related target domain. This approach does not consider missing modality conditions of co-learning.

Instructional videos and text instructions of cooking recipes are readily available on the web. The sequence of instructional steps in the two recipes is different among the text and video recipes but are the same for a cuisine dish. An unsupervised alignment algorithm [86] is used to create alignments between two text recipes, two video recipes, text and video recipes. One method is to create an alignment between text and video by mapping between words in sentences to objects in a video. Another approach uses text instructions and transcripts of a video to create mapping based on word similarity as words offer better semantic information than videos. The latter approach is used to create alignment between text-text, video-video, and text-video. With this approach, performance better than the current benchmark is obtained for the human-annotated YouCook2 [110] dataset. Temporal relationship in unlabeled videos is used by creating text-video and audio-video pairs to train multimodal transformers using MIL-NCE [111] for image classification, video action recognition, audio event classification, and zero-shot video retrieval. The transformers share weights among the text, audio, video modalities creating a general-purpose model for all modalities with separate tokenization and linear projection.

Thus, creating an alignment using unsupervised techniques can facilitate co-learning. Alignment among the modalities is crucial for the transfer of knowledge. The methods like contrastive learning are showing promising results for multimodal tasks.

#### *4.4.5 Discussion*

The supervised multimodal models using labeled data support co-learning objectives of missing or noisy modality. However, annotation is costly. It is a time-consuming manual process that is error-prone. It demands expertise, and the complexity increases for multimodal data. Along with annotation, alignment among modalities and noise also needs to be ensured. Hence, it is essential to have techniques that can achieve co-learning without labeling. Creating semantic embedding spaces with available labeled data and using it for multimodal tasks with only one modality at test time is possible using semi-supervised learning. The pre-trained models like object detectors and sentiment classifiers can create labels for one modality and the inherent relationship among modalities is used for creating multimodal labelled data. This approach is promising for co-learning with semi-supervised and un-supervised learning.

The summary of key co-learning studies which use semi-supervised, unsupervised, and weakly supervised techniques is presented in Table 5.

**Table 5**: Summary of studies for modality annotation objective of co-learning

| Modality Annotations | Sub-class | Multimodal Taxonomy Area | Summary of the studies | | | Observations and features | References |
|---|---|---|---|---|---|---|---|
| | | | Modalities | Applications | Methods | | |
| Supervised | All modalities are | Representation, Fusion, Co-learning | Audio, Video, Text | Sentiment Analysis | Memory Fusion Network | Manual labeling and tools for alignment and feature extraction. | [15] |
| | labeled | Representation, Co-learning | Image, Text | Vision-language Tasks | Multi-task learning | Language is person-dependent; universal ground truth is difficult even for supervised data. | [17] |
| Semi-supervised | One modality labeled other is unlabeled | Representation, Alignment, Co-learning | Image, Text | Vision-language Tasks | Multimodal Embedding | Use of un-annotated text for a semantic relationship with image labels forming embedded space. | [83] |
| Un-supervised | NA | Representation, Alignment, Fusion, | Image, Text | Phrase Grounding | Contrastive Learning, fusion-attention | Object detectors create a mapping between image regions and phrases in the caption. | [22] |
| | | Representation, Alignment | Audio, Video | Deception Detection | Transfer Learning, Sub-space Alignment | Mapping between high stake deception using supervised data from low stake deceptions. | [109] |
| | | Representation, Alignment Fusion, Co-learning | Audio, Video, Text | Video retrieval using a text query | Transformers MIL-NCE | Temporal relationship within an unlabeled video is used to create alignment | [111] |
| | | Representation, Alignment | Video, Text | Alignment Task | Component Level Alignment (video objects – recipe text) | Unsupervised alignment using overlap among different modality instruction steps. Graph network is created for each dish. | [86] |
| Weakly Supervised | Concept Level | Representation, Alignment, Fusion, | Image, Text | Phrase Grounding | Contrastive Learning, fusion - attention | Low-level phrase-objects labels created using high-level (weak) image captions. | [22] |
| | | Representation, Alignment, Co-learning | Image-Text | Vision-Language Tasks | Multimodal Embedding | Visual co-occurrences between objects and attribute words derived from image-caption labels. | [108] |

*4.5 Multimodal domain adaptation*

Having labeled data for the target task sometimes is challenging; instead, we get labeled data from a different dataset but related to the domain of a target task. *Domain adaptation* (DA) [58] creates models for a target domain by training on labeled samples from the source domain and leveraging unlabeled samples from the target domain as supplementary information during training. There may be few labeled target samples at training time for *few-shot learning* methods. In *domain generalization* (DG) [112], labeled data is available from multiple source domains, but no target domain samples are available. Both DA and DG have common label space in source and target domains.

Domain adaptation is further classified as *homogenous* and *heterogeneous DA* [113]. The source and target domains have similar or the same dimensionality feature spaces in homogenous DA and different feature spaces in heterogeneous DA. Domain generalization can also be classified as *homogenous* and *heterogeneous DG* [114]. In homogenous DG, label data is not available from the target domain, but the target label space is the same as the source label space. Whereas in heterogeneous DG, label spaces are different or disjoint at the source and target [115]. Domain adaptation is classified as *Supervised, Unsupervised, and Semi-supervised* based on the availability of labels for the source domain.

*Unimodal DA* focuses on domain shift between source and target domain, e.g., source dataset is MNIST, and the target dataset is SVHN or source is one type of product reviews, and target is another type of product reviews. In the case of *homogenous multimodal DA*, dimension of modality gets added, e.g., if the source domain audio and video modality dataset recorded in the controlled conditions, and the target dataset is collected in the wild conditions; hence there is a difference of domain for background, scenes, sound conditions, etc. For multimodal DA, each modality has different structures, content and may have different domain shifts. In heterogeneous multimodal DA, source and target domains have distinct representations with different modalities, such as image and text.

#### *4.5.1 Homogenous multimodal domain adaptation*

Homogenous multimodal DA has multiple modalities at source and target domain instead of one modality at both. For example, image and text in Flickr dataset as a source and image and text in Google News as target domain having different noisy conditions, i.e., domain shift.

Intra-modality, inter-modality, and inter-domain properties are used for multimodal DA with video and audio modality for emotion recognition and image and text for cross-modal retrieval tasks [116]. The domain invariant features of each modality are learned using an adversarial network on a modality from the source domain and the same modality from the target domain. The objective is to train a model such that fused output features cannot discriminate between the source and target, removing multimodal domain shift.

In the application of social media event rumor detection [117], social post is disentangled into content-specific information and rumor style specific information. This approach helps to train a classifier on rumor-specific information rather than content information to meet the objective of rumor detection irrespective of its content, as rumor can occur for different types of events. The adversarial domain adaptation method is used to tackle the lack of labeled data in the new event. Here, the missing modality condition of co-learning is not achieved, but the source domain dataset helping a task on the target domain is achieved. For emotion recognition [118], cross-corpora, cross-language, cross-speaker, cross-modality, and emotion elicitation are considered domain shifts. Adversarial training is employed to capture emotion-related information and remove domain-specific information to achieve domain adaptation.

#### *4.5.2 Heterogeneous multimodal domain adaptation*

In unimodal heterogeneous DA, there are different feature spaces in the source and target domain. This definition can be extended to multimodal when the source and target domains have different modalities.

Transfer learning utilizes knowledge from the source domain to predict tasks on an unknown target domain. In multimodal transfer learning, it is assumed that both source and target modalities are available at training time. However, that's not always true, and there may be a missing modality in the target domain. This scenario is classified as heterogeneous multimodal DA. The challenge is addressed by having an auxiliary database with complete modalities [119]. Assuming that if one modality can transfer knowledge to another in an auxiliary database and there is a knowledge transfer between databases, then the available modality in the main database can transfer knowledge to the missing modality. Here, knowledge transfer is in two directions - cross-modality and cross-database using a latent low-rank constraint on subspace. It is implemented with various modalities of an image that can be extended to modalities like image-text.

Heterogeneous DA is related to multi-view learning. Multiple views provide better representation and can work when both views are not available. This concept is used to solve heterogeneous DA problems using multi-view auxiliary data, which relates two views that otherwise are different in source and target. This auxiliary data could be co-occurrence data or embedding spaces formed by projecting source and target domain data samples. The projection of each domain in domain invariant latent space is called symmetric transformation. A transformation from source domain space to target domain space is called asymmetric transformation-based heterogeneous DA [113]. Some of the latest deep DA methods are based on Siamese architecture, having two streams for source and target domains with classification loss and discrepancy or adversarial loss. Discrepancy loss works to reduce the domain shift, whereas adversarial loss forms a common space. Domain translator function learned from multimodal source data can predict class even if one modality is absent [120] using weakly shared deep transfer networks for domain adaptation.

For multimodal sentiment analysis, text modality is used as a source domain and audio-video modalities as a target domain [70], forming a heterogeneous DA set-up. Text (source) modality representation is used to reconstruct audio-video modality (target) representations using a decoder to correlate. GAN generates the samples for the target domain for audio-visual cross-modal mapping in [121] dacssGAN (Domain Adaptation Conditional Semi-Supervised Generative Adversarial Network). The baseline approach would be to provide noise and samples of a dataset, having target domain distribution as an input to the generator to generate samples of the target domain discriminated by the discriminator for genuineness.

Thus, heterogeneous multimodal DA can be achieved with auxiliary datasets, embedding spaces using co-occurrences, transfer learning approach or GANs.

#### *4.5.3 Homogenous multimodal domain generalization*

Homogenous multimodal DG is defined as labeled multimodal data is available from multiple source domains, but no target domain multimodal samples are available. The number of modalities can be the same or different at source and target, but the target label space is the *same* as the source label space. Deep transfer learning with score fusion [122] predicts emotions on wild emotion data when trained on acted emotion data having audio and video modality. For abnormal gait recognition [123], motion features, RGB-D, and electromyography (EMG) data are employed with an autoencoder. Different person's data is used during training and testing to achieve cross-subject cross-modal domain generalization.

### 4.5.4 *Heterogeneous multimodal domain generalization*

Heterogeneous multimodal DG is employed when labeled multimodal data is available from multiple source domains, but no target domain samples are available. The number of modalities can be the same or different at source and target, but label spaces are *different* at the source and target.

Cross-modal generalization is defined when the model has one modality at source and another modality at target and can perform a new task in a target domain [95]. For example, the source is the image, and the target is an audio, and the objective is to align shared knowledge from image to audio. The generalization won't need samples or labels from the target domain except for few samples for few-shot learning. Here cross-modal meta-alignment is proposed to create a space where representations of similar concepts in different modalities are closer while ensuring generalization to new tasks using a classifier. CROMA (Cross-modal Meta-Alignment) algorithm is proposed in which alignment is done by using Noise Contrastive Estimation (NCE) instead of using a translation model with Maximum Likelihood Error (MLE) while mapping. Meta-alignment trains encoder on the source and target modalities across multiple alignment tasks. Model is evaluated for text-image, image-audio, text-speech and performed better than meta-learning and domain adaptation models.

In cross-modal data programming [124], labeling functions are used on an auxiliary dataset of clinical reports and medical imaging instead of manually labeling any part of data. In this weak-supervised approach, medical images are used as target modality at test time. Multi-domain and Multi-modality Event Dataset (MMED) [125] is released to enable domain generalization for cross-modal retrieval. Text data from different news portals and associated images from social media are utilized in a weakly supervised manner. Cross-domain cross-modality transfer learning [126] exploiting dictionary base alignment is proposed for cross-modal retrieval.

### 4.5.5 *Discussion*

Domain adaptation has gained a lot of focus recently as it takes deep learning more towards real-life conditions where we have different conditions at inference time than at training time. There is significant progress for unimodal applications, and efforts are on for multimodal DA. The definitions and classifications of multimodal DA are still evolving. We reviewed available unimodal literature for definition and classification and searched research papers for multimodal DA applications. The objective is to highlight how co-learning among the modalities can achieve DA.

Domain adaptation is a step to achieve co-learning objectives as it supports modalities having different distributions in source and target domains and also missing modalities at the target domain for heterogenous DA. Intra-modality and inter-modality attributes, along with inter-domain attributes, are used to arrive at domain-invariant features to address domain shift. Also, auxiliary databases enable knowledge transfer from a modality of one database to a modality of another database to achieve the missing modality objective of co-learning. Domain adaptation, along with co-learning among the modalities, is a promising technique to address real-life conditions. The multimodal domain adaptation studies presented are summarized in Table 6.

**Table 6**: Summary of studies for domain adaptation objective of co-learning

| | | | Summary of the studies | | |
|---|---|---|---|---|---|

| Adaptation Type | Sub-class | Multimodal Taxonomy Area | Modalities | Applications | Methods | Observations and features | References |
|---|---|---|---|---|---|---|---|
| Multimodal Domain Adaptation (DA) | Homogenous Multimodal DA | Representation, Fusion | a) Audio, Video b) Image-Text | Emotion Recognition, Cross-Media Retrieval | Covariant attention with adversarial learning | Constraints at intra and inter-modality and at prediction to learn discriminative and domain adaptive features. | [116] |
| | | Representation | Image, Text | Social Media Rumor Detection | Multimodal disentanglement | Social media post is disentangled into content and rumor specific information. | [117] |
| | | Representation | Audio, Video | Emotion Recognition | Adversarial learning | Emotion elicitation is considered as domain shift | [118] |
| | Heterogeneous Multimodal DA | Representation, Co-learning | Image | Face Recognition, Image Classification | Transfer learning (Heterogeneous) | An auxiliary database with complete modalities used for cross-modality and cross-database knowledge transfer. | [119] |
| | | Representation, Co-learning | Audio, Video | Emotion Prediction | GAN (dacssGAN) | Video used to generate audio target using cGAN and class labels. Conformal prediction is used for semi-supervised learning. | [121] |
| Domain Generalization (DG) | Homogenous DG | Representation, Fusion | Audio, Video | Emotion Prediction | Transfer learning | Trained on acted emotion dataset and tested on wild emotion dataset | [122] |
| | | Representation | Motion, EMG, RGBD | Gait Prediction | Multimodal autoencoder | Cross-subject cross-modal knowledge transfer for domain generalization. | [123] |
| | Heterogeneous DG | Representation, Co-learning | Image, Audio | Audio Classification | Cross-Modal Alignment – Meta-Learning | Meta-alignment trains encoder on the source and target modalities across multiple alignment tasks. | [95] |
| | | Representation, Co-learning | Text, Image | Medical Image classification | Cross-modal retrieval | Cross-modal data programming using an auxiliary dataset in a weak supervision. | [124] |
| | | Representation, Co-learning | Text, Image | Media Event Classification | Cross-modal retrieval | Data from different sources combined with weak supervision | [125] |
| | | Representation, Alignment, Co-learning | Text, Image | Cross-modal retrieval | Transfer Learning | Dictionary-based alignment with cross-domain cross-modality transfer learning | [126] |

### 4.6 *Interpretability and fairness*

The objectives like interpretability, explainability, fairness, and bias are emerging research areas and are very relevant in multimodal deep learning applications. We specifically included and discussed them in the co-learning taxonomy. The researchers should ensure the interpretability and fairness objectives along with co-learning objectives. The goal is to understand if multimodality at training time improves interpretability and fairness for unimodality at test time.

#### 4.6.1 *Interpretability and explainability*

Machine learning marks widespread adoption across many real-world applications, including high risk and critical applications like healthcare, finance, regulatory compliance, and human-machine interaction [127]. With the recent advancement in deep learning, accuracies have increased. Still, their results are less interpretable [128] as models are inherently complex and have deep hidden layers to establish a non-linear relationship. Hence there is a need for users to understand results provided by machine learning models. Models are interpretable when it is evident why the decisions are made and explainable when it is evident how the system made the decisions [129].

Multiple survey studies presented interpretable machine learning classification, challenges, and methods [130,131]; however, many of these studies are focused on unimodal model interpretation. Multiple modalities provide complementary information, which is leveraged for better explanations. It is showed that multimodal models designed for explainability are robust towards adversarial attacks and less biased. Multimodal explainability, challenges, methods to achieve explainability, datasets, and tools are covered in [132]. In this paper, we highlight the relation between multimodal co-learning and interpretability and explainability, i.e., does multimodal co-learning help achieve interpretability and explainability, or is there an impact on interpretability and explainability by adding a co-learning mechanism. Also, can recent methods like disentanglement, multimodal factorization, contrastive learning, and counterfactuality, which are used for co-learning implementation, help achieve interpretability and explainability.

Multimodal models are more interpretable and explainable [132] than unimodal ones. Multimodal co-learning aids in interpreting multimodal model results; it shows how each modality contributes to overall prediction or which are shared factors and modality-specific factors among the modalities. *Multimodal Factorization Model* (MFM) [67] factorizes multimodal data into a) discriminative factors, which are common across all modalities, and b) generative factors, which are specific to modality. Information-based interpretation over the entire dataset and gradient-based interpretation over a video segment is performed to study the contribution of each modality, and results are the same as human observations attributing to either presence of specific words or facial expressions.

The *disentanglement* based on fashion attributes such as color, style, etc., creates representations that can correspond to specific features rather than creating a common representation in space for image and text [133]. This implementation based on the autoencoder approach helps deal with missing and noisy modalities for cross-modal image retrieval using natural language queries with higher interpretability.

Cross-Modal Cycle Generative Adversarial Network (CMCGAN) [75] helps to generate any of the absent modality, label noise-robust GAN (rGAN) [94] using Conditional GAN (cGAN) and auxiliary classifier GAN (AC-GAN) can handle noisy labels, and dacssGAN (Domain Adaptation Conditional Semi-Supervised Generative Adversarial Network) is used to generate audio modality using visual modality [121] under multimodal domain adaptation set up. By maximizing mutual information between noise variables and observation, GAN produces interpretable representation with InfoGAN [134]. It indicates that GAN-based architecture can help both co-learning as well as interpretability.

*Counterfactual explanations* are based on conditions of observations; for example, certain observations are present or missing leading to a specific output. Multimodal co-learning implementation inherently supports working with missing modalities partly or fully and various noisy conditions. Making text input absent in VQA models, bias towards language modality is studied using a counterfactual approach [135]. Similarly, visual biases are studied by making visual input absent or distorted or replaced by wrong ones using a counterfactual approach [136]. It highlights that multimodal co-learning, which does model work when the modality is missing, removes bias towards a modality and makes models more interpretable. In the phrase grounding, the association between an image region and a phrase from a caption is created for downstream tasks like VQA, image captioning, and image-text retrieval. This approach allows the model to use weak supervision at the image and caption level and increases the interpretability of the model's output [137]. Adding *contrastive learning* in phrase grounding setup [108] to maximize mapping between regions and corresponding captions compared to non-related regions and captions improves the accuracy and interpretability.

Thus, multimodal co-learning enables the interpretability and explainability of multimodal models via multimodal factorization, disentanglement, variants of GANs, counterfactual explanations, and contrastive learning methods.

#### *4.6.2 Fairness and bias*

Training data, algorithms, choice of output classes all contribute to bias and impact the fairness of machine learning models. Even the choice of output class should be well thought else model tries to show some output forcefully, which could be biased. Other sources of bias are annotators' personal biases and affiliations, imbalanced data, and feature selections. The taxonomy of bias in machine learning has been classified into 22 different categories the biases [138]. In multimodal data, these sources of bias can impact multiple modalities, impacting model predictions. Each modality has unique representations and contributions in the decision-making process, thereby making models biased towards a particular modality while ignoring the contribution of other modalities.

In VQA tasks, language often becomes the dominant modality and neglects clues from the visual modality. It impacts the performance of the model when unseen data is provided during inference [139]. Also, the model creates a bias towards a dataset, thereby failing to generalize cross-dataset for a VQA task [140]. For detecting bias in the recruitment process, Fair Automatic Recruitment (FairCVtest) [141] testbed is designed using 24,000 synthetic resumes and made available to the public for further research. The testbed showed that bias towards gender and ethnicity in the recruitment process exists even when certain information is masked; this could be very harmful to the recruitment process.

Recently, Contrastive Language-Image Pre-training (CLIP) [142], in spite of using a pre-training image-captioning on 400 million image-text pairs of data, has a social bias in the model. They designed probes to study bias in the model by focusing on geography, social conditions, race, color, age, etc. Model is biased in many conditions, e.g., more men images are classified as crime-related than women, terrorism got biased towards Asians and immigration towards Latin America. CLIP is evaluated on FairFace [143] dataset. MANGO [103], which adds the noise in the embedding space of VL models, helps reducing bias at the embedding layer. Additionally, some regions of the input images are masked, and some tokens of text are dropped which, further helps to remove the bias towards data samples.

#### *4.6.3 Discussion*

Multimodal co-learning objective to handle missing and noisy helps to remove bias caused due to various reasons. The missing modality helps to interpret multimodal model results as it shows how each modality contributes to overall prediction. Multimodal factorization and attributes disentanglement support missing modality along with interpretability of results. GAN frameworks that generate missing and noisy modalities also increase interpretability using its discriminative property. Thus, experiments to ascertain co-learning objectives can also help to understand interpretability and bias multimodal systems. The summary of studies for interpretability and fairness is presented in Table 7.

**Table 7**: Summary of studies for interpretability and fairness objective

| Interpretability and Explainability | Sub-class | Multimodal Challenge area | Summary of the studies | | | Observations and features | References |
|---|---|---|---|---|---|---|---|
| | | | Modalities | Applications | Methods | | |
| Interpretability and Explainability | NA | Representation, Fusion, Co-learning | Audio, Video, Text | Sentiment Analysis | Multimodal Factorization | Interpretation using information and gradient-based methods to know contributions of individual factors towards prediction. | [67] |
| | | Representation, Fusion, Co-learning | Image, Text | Cross Media Retrieval | Multimodal Disentanglement | Disentanglement of fashion attributes using encoder-decoder and increasing interpretability. | [133] |
| | | Representation | Image | Image Classification | Disentangled Representation | Provides interpretable representations - style & digit shape for MNIST, and eyeglasses and emotions on CelebA dataset using GAN. | [134] |
| | | Representation | Image, Text | Visual Question Answering | Counterfactual generation | Counterfactual images change the image semantics and the output of VQA shows internal mechanism models. | [136] |
| | | Representation, Fusion | Image, Text | Image Phrase Grounding | CNN, Attention, Counterfactuals | Decomposes text into the entity, semantic attribute, color, etc. | [137] |
| Fairness and Bias | NA | Representation, Fusion | Image, Text | Recruitment | CNN with Fusion on Structured and | Fair-Cvtest bed is open-sourced. Users can select features to check gender and | [141] |
| | | | | | Unstructured data | ethnicity bias varying bias levels in the model. | |
| | NA | Representation, Fusion, Co-learning | Image, Text | Object Classification, Action recognition, OCR | Multimodal Embedding with Contrastive Learning | CLIP pre-trained on 400 million image-text pairs of data still got a social bias. The probes were designed to detect bias. FairFace dataset is used as a benchmark. | [142] |

**NA – Not Applicable*

### *4.7 Summary of multimodal co-learning taxonomy*

Throughout this section, we discussed key research studies which implemented one or more objectives of multimodal co-learning. We also included some studies which don't have any missing modality at test time but achieved other co-learning objectives. We created the consolidated view of studies discussed for each co-learning objective in Table 8 and included corresponding multimodal challenge areas (as discussed in Section 1.1). This consolidated view helps to understand which co-learning objectives are widely addressed and which need further focus in the future.

**Table 8**: Summary of key studies and co-learning objectives achieved by them

| Sr. No | Modalities at Training | Modalities at Testing | Multimodal Taxonomy Challenges | | | | | Presence of Modality | | | | Data Parallelism | | | Noisy Modality | | Modality Annotations | | | | Domain adaptation | | Interpretability & Fairness | | References |
|---|---|---|---|---|---|---|---|---|---|---|---|---|---|---|---|---|---|---|---|---|---|---|---|---|---|
| | | | Representation | Alignment | Translation | Fusion | Co-learning | Fully Missing | | Partly Missing | | Parallel | Non-Parallel | Hybrid | Data Noise | Label Noise | Supervised | Semi- Supervised | Un-supervised | Weakly supervised | Domain Adaptation | Domain Gen.² | Inter. & Explain.³ | Fairness & Bias | |
| | | | | | | | | Train | Test | Train | Test | | | | | | | | | | | | | | |
| 1 | Image, Text | Image | ■ | | | | ■ | | ■ | | | ■ | | | | ■ | | ■ | | ■ | | | | | [48] |
| 2 | Audio, Video, Text | Audio, Video | ■ | | | | ■ | | ■ | | | ■ | | | | | | | | ■ | | | | | [47] |
| 3 | Audio, Video | Audio | ■ | | | | ■ | | ■ | | | | ■ | | | | | | | ■ | | | | | [72] |
| 4 | SAR, Optical Image | Optical Image | ■ | | | | ■ | | ■ | | | ■ | | | | | ■ | | | | | | | | [9] |
| 5 | RGB, Depth | RGB | ■ | | | | ■ | | ■ | | | ■ | | | ■ | | ■ | | | | | | | | [65] |
| 6 | Audio, Video | Audio or Video | ■ | | | ■ | ■ | | ■ | | | ■ | | | | | ■ | | | | | | | | [24] |
| 7 | Audio, Video, Text | Audio-Video, Audio-Text, Video-Text | ■ | | | ■ | ■ | | ■ | | | ■ | | | | | ■ | | | | | | | | [67] |
| 8 | Image, Text | Image or Text or Image-Text | ■ | ■ | | | ■ | | ■ | | | ■ | | | | | ■ | | | | | | | | [68] |
| 9 | Audio, Video | Audio or Video | ■ | ■ | ■ | | ■ | | ■ | | | ■ | | | | | ■ | | | | | | | | [21] |
| 10 | Image, Text | Image or Text or Image-Text | ■ | | | | ■ | | ■ | | | ■ | | | | | ■ | | | | ■ | | | | [17] |
| 11 | Image, Text | Image or Text | ■ | | | ■ | ■ | ■ | ■ | | | ■ | | | | | ■ | | | | | | | | [6] |
| 12 | Audio, Video, Text | Text | ■ | | | ■ | ■ | | ■ | | | ■ | | | | | ■ | | | | ■ | | | | [15] |
| 13 | Audio, Video, Text | Text | ■ | ■ | ■ | | ■ | | ■ | | | ■ | | | | | ■ | | | | | | | | [23] |
| 14 | Audio, Video | Audio | ■ | | | ■ | ■ | | ■ | ■ | | ■ | | | ■ | | ■ | | | | | | | | [77] |
| 15 | Image-Text Audio-Image | Text or Image | ■ | | | | ■ | | ■ | ■ | ■ | ■ | | | | | | | | | | | | | [59] |
| 16 | Image, Text | Image | ■ | | | | ■ | | ■ | | | | ■ | | | | | ■ | | | | | | | [83] |
| 17 | Video, Text | Text | ■ | | | | ■ | | ■ | | | | ■ | | | ■ | | | | ■ | | | | | [79] |
| 18 | Image, Text | Text | ■ | ■ | | | ■ | | ■ | | | | | ■ | | | ■ | | | | | | | | [10] |
| 19 | Video, Text | Video, Text | ■ | ■ | | | | | | | | | | ■ | | | | | ■ | | | | | | [86] |
| 20 | Video, Text | Text | ■ | ■ | | | ■ | | ■ | | | | | ■ | | | | | | ■ | | | | | [84] |
| 21 | Image, Text | Image or Text | ■ | | ■ | | ■ | | ■ | | | | | ■ | | | ■ | | | | | | | | [85] |
| 22 | Audio, Ontology | Audio | ■ | | | ■ | ■ | | ■ | | | | | ■ | | ■ | | ■ | | | | | | | [91] |
| 23 | Image | Image | ■ | | | | | | | | | ■ | | | | ■ | ■ | | | | | | | | [94] |
| 24 | Image | Audio | ■ | ■ | | | ■ | | ■ | | | ■ | ■ | | | ■ | ■ | | | | | ■ | | ■ | [95] |
| 25 | Image, Text | Image, Text | ■ | | | | | | | | | ■ | | | ■ | | ■ | | | | | | | | [103] |

| | | | | | | | | | | | | | | | | | | | | | | | | |
|---|---|---|---|---|---|---|---|---|---|---|---|---|---|---|---|---|---|---|---|---|---|---|---|---|
| 26 | Audio, Video | Audio | ■ | | | ■ | ■ | | | | | ■ | | | ■ | | | | | ■ | | | | | [102] |
| 27 | Text, Image, HTML layout | Text, Image, HTML layout | ■ | | | ■ | | | | | | ■ | | | | | ■ | | | | | | | | [106] |
| 28 | Text, Image | Text, Image | ■ | ■ | | | | | | | | | ■ | | | | | | ■ | ■ | | | | | [22] |
| 29 | Text, Image | Text | ■ | ■ | | | ■ | | ■ | | | | ■ | | | | | | | ■ | | | | | [108] |
| 30 | Audio, Video | Audio, Video | ■ | ■ | | ■ | | | | | | ■ | | | | | | ■ | ■ | | ■ | | | | [109] |
| 31 | a) Audio, Video b) Image, Text | a) Audio, Video b) Image, Text | ■ | ■ | | ■ | ■ | | ■ | | | ■ | | | | | ■ | | | | ■ | | | | [116] |
| 32 | Image, Text | Image, Text | ■ | | | | | | | | | ■ | | | | | ■ | | | ■ | ■ | | | | [117] |
| 33 | Image | Image - different domain | ■ | ■ | | | | | ■ | | | ■ | | | | | ■ | | | | ■ | | | | [119] |
| 34 | Video | Audio | ■ | | | | ■ | | ■ | | | ■ | | | | | ■ | ■ | | | ■ | | | | [121] |
| 35 | Image, Text | Image or Text | ■ | | | ■ | ■ | | ■ | | | ■ | | | ■ | | ■ | | | | | | ■ | | [133] |
| 36 | Image | Image | ■ | | | | | | | | | | | | | | | | ■ | | | | ■ | | [134] |
| 37 | Image, Text | Image, Text | ■ | | | | | | | | | ■ | | | | | | ■ | | | | | ■ | | [136] |
| 38 | Image, Text | Image, Text | ■ | | | ■ | | | | | | | ■ | | | | | | | ■ | | | ■ | | [137] |
| 39 | Image, Text | Image, Text | ■ | | | ■ | | | | | | ■ | | | | | ■ | | | | | | | ■ | [141] |
| 40 | Image, Text | Image | ■ | | | | ■ | | ■ | | | | ■ | | | | | | ■ | | | | | ■ | [142] |
| 41 | Audio, Video, Text | Audio, Video | ■ | ■ | | ■ | ■ | | ■ | | | | ■ | | | | ■ | | | | ■ | | | | [70] |
| 42 | Audio, Image | Audio or Image | ■ | | | | ■ | | ■ | | | ■ | | | ■ | | ■ | | | | | | | | [75] |
| 43 | Audio, Video, Text | Text | ■ | | | ■ | ■ | | ■ | | | | | ■ | | | ■ | | | | ■ | | | | [87] |
| 44 | EEG, Image | EEG | ■ | | | ■ | ■ | | ■ | ■ | | ■ | | | | | | ■ | | | | | | | [11] |
| 45 | Audio, Video, Text | Text | ■ | ■ | | ■ | ■ | | ■ | | | | ■ | | ■ | | | | ■ | | | | | | [111] |
| 46 | Image, Text | Image | ■ | ■ | | | ■ | | ■ | | | ■ | | | | | | | | ■ | | ■ | | | [124] |
| 47 | Motion, RGB-D, EMG | Motion, RGB-D, EMG | ■ | ■ | | | | | | | | ■ | | | ■ | | ■ | | | | | ■ | | | [123] |
| 48 | Audio, Video | Audio, Video | ■ | | | ■ | | | | | | ■ | | | | | | | | | ■ | | | | [118] |
| 49 | Image, Text | Image, Text | ■ | ■ | | | ■ | | ■ | | | | | ■ | | | | | | ■ | | ■ | | | [126] |
| 50 | Audio, Video | Audio, Video | ■ | | | ■ | | | | | | ■ | | | ■ | | ■ | | | | | ■ | | | [122] |

[2] Domain Generalization
[3] Interpretability and Explainability

## 5. Computational approaches for multimodal co-learning

Implementation of co-learning objectives and sub-objectives can be achieved using multiple multimodal deep learning techniques, as discussed in Section 4. Until a few decades ago, the multimodal problem was treated as a multi-view learning problem with co-teaching, teacher-student network, co-training, etc., methods. Recently the implementation has focused on using different types of fusion techniques. However, all five areas, namely, representation, alignment, translation, fusion, and co-learning, are being used to address tasks. Specifically, co-learning, which supports real-life conditions, is an emerging area of high importance in multimodal deep learning.

This section elaborates on some of these latest computational approaches and how they support the co-learning objectives.

### *5.1 Fusion*

The first step to designing multimodal systems is to take two or more modalities and fuse them to solve downstream tasks. The fusion techniques are classified as early (feature), late (decision), or intermediate (hybrid) fusion [13], depending on the level in the network at which representations are fused. Fusion is still specific to the data, domain, and task at hand, and there are no generalized fusion rules. Early fusion cannot consider intra-modality specifics, and late fusion cannot capture inter-modality specifics; hence, hybrid fusion is more commonly used. Hybrid fusion provides the ability to fuse data at the required level and obtains better joint representation in the combined feature space. The modalities are fused using concatenation, multiplication, or weighted sum.

Several researchers have reported various fusion techniques to facilitate better representation. Tensor fusion [27] captures intra and inter-modality features better than early, late, and hybrid fusion techniques, forming a large representation space. Low-rank multimodal fusion [26] reduced the representation dimensions and computation complexity. The approach to combine two modalities (e.g., text, video) first and then to add the third modality (e.g., audio) to learn bimodal and trimodal correlations is termed as a hierarchical fusion [144]. Multimodal fusion architecture search space [145] is used to decide which layers to use for fusion from each modality and which non-linear function to be used for fusion. Heterogeneous representations of modalities need different learning rates and optimization strategies. The same optimization strategy for all modalities leads to overfitting. Hence, the overfitting-to-generalization ratio [146] is used to control loss weighing in the training phase.

Late fusion can manage missing modalities at test time but is often underperforms compared to intermediate fusion. Hence hybrid fusion, multi-task learning, memory networks, and attention mechanisms are utilized for missing and noisy modalities conditions. Memory fusion network [15] creates view-specific and cross-view-specific interactions to handle missing modalities. Audio, video, and text modalities are used during training, whereas only text modality is used during testing, proving that co-learning increases performance compared to unimodal text. The gap between fusion and representation learning is closing. Both are used with other multimodal concepts such as alignment and translation to implement multimodal co-learning effectively.

### *5.2 Attention and transformers*

The sequence-to-sequence models with encoders like LSTM [147] can process variable-length input sequences and generate output sequences for language tasks. However, sequence-to-sequence models cannot handle long sequences [49].

**Attention** focuses on certain aspects of information, like specific features, regions in an image, or a time step in the sequence. Attention can be *soft* (attention is obtained by looking at all features or image regions) or *hard attention* (attention is obtained by looking at pre-determined regions or features) [148]. Soft attention is preferred in practice because of its ease of optimization. Based on the context used for attention, those are also classified as *local* and *global* attention. The attention used on information in the same modality to capture the temporal relationship and obtain better representation is called *self-attention*. *Graph-based attention* [149] uses domain knowledge to extract the relationship between different entities of modalities. Using individual attention at the sentence and word levels and passing information from lower level to higher levels as *hierarchical attention* [150] is effective in a document classification task. *Co-attention* [81] is used to learn the correlation between two paired entities like two documents or paired modality data samples. Attention has been the important step to create alignment between modalities instead of creating alignment manually. In multimodal caption generation or phrase grounding models, attention establishes the mapping between words and image regions, thereby explaining how the model makes a prediction.

**Co-attention** is used to create alignment between different modalities; for example, words in a question are mapped to objects in the image for VQA tasks [80]. Co-attention enables multimodal co-learning by creating alignment among the modalities. *Hierarchical co-attention* [81] is designed to focus at a word, phrase, and sentence level between question and image and then recursively combined from word to sentence level in VQA task. The co-attention encoder module [151] correlates latent features of each view in GAN to provide weightage to input features. It provides interpretability as weightage to each view is known in the multi-view subspace learning space. The weighing scheme is extended to noisy and good features, thereby increasing robustness to noisy samples. *Joint co-attention* [152] employs joint representation of audio and video to create co-attention with individual modalities instead of co-attention between two modalities. It improved accuracy for audio-visual event localization even in the presence of noisy inputs.

**Transformers** consist of two networks, an encoder and decoder, and an attention mechanism to capture temporal information. These can perform parallel computation and handle large datasets compared to sequence-to-sequence [147] LSTM or RNN models. The use of transformers for the language domain outperformed many state-of-the-art results and generated huge research interest to adapt it for multiple modalities. *LXMERT* [153] consists of object relationship encoder, language encoder, and cross-modality transformer encoders. It is trained on five different tasks that enable the model to learn features from the same modality or aligned elements in other modalities. Bidirectional representation learning is added to the transformer to get pre-trained model *BERT* [51], which can be fine-tuned for specific tasks. It is extended to *mBERT* by training on 104 languages from wiki using masked language modeling to have a multilingual capability in the transformer. The BERT models extended to VideoBERT [154], ViLBERT [155], and VisualBERT [156] for multimodal tasks like video captioning, image captioning, VQA, etc. Text-video and audio-video pairs from unlabeled videos are used to train multimodal transformers using MIL-NCE [111].

Three transformers were employed for the emotion and sentiment recognition task [157], one for each modality – text, audio, and video. Joint encoding with co-attention module helped to achieve state-of-the-art performance. The transformer is modified to process multimodal input. The addition of co-attention block [158] followed by fusion helps to select and use required features based on the similarity between the features of different modalities providing better accuracy of the image-text classification task. The multilingual transformer has been extended to handle video modality [84] for video search using multilingual text. Video modality is considered as pivot modality, which got an inherent relationship with multilingual captions.

Thus, attention, co-attention, and transformers enable multimodal co-learning to manage missing modality, noisy conditions, data parallelism, and data annotations (unsupervised).

### 5.3 Encoder-decoders

In the **Encoder-decoder** framework, the encoder encodes a source sentence into a fixed-length state, from which a decoder translates into a variable-length sentence. The encoder-decoder framework is extended for multimodal data in which input is a source modality and output is a target modality. CNN [159] as encoder and RNN [147] as decoder with visual attention is used for image captioning [148] and to generate video description [160]. The latent vector can capture shared information between source and target as an error is flown from output to input during training. Attention mechanism helps as it focuses on intermediate representations between source and targets and selects important features, avoiding loss of information. An encoder [161], with self-attention on features of questions and a decoder with question-guided attention on the image and self-attention on an image, creates features of an image to provide better accuracy on VQA task. Encoder-decoder helps generate target modality representation based on the source modality representation, but the application of the encoder-decoder mechanism for more than two modalities is complex.

**Autoencoder** is the same as encoder-decoder, but input and output are the same, i.e., decoder reconstructs input to minimize reconstruction loss. The denoising autoencoder works when some of the values are missing, and robust encoders modify the loss function to get robustness against noisy conditions [162]. Autoencoder is extended to use for multimodal data and is called *multimodal autoencoder* [163]. Two encoder-decoder modules are used separately but combined with having a shared representation. Each encoder-decoder module handles one modality, i.e., audio and video, and the decoder reconstructs these modalities using two decoders to minimize the reconstruction loss.

Correspondence Autoencoder (Corr-AE) [164] does representation and correlation-learning together. Reconstruction loss and the similarity between representations of two modalities are used to perform the cross-modal retrieval task of image-text. Thus, autoencoders can extract shared information from latent space and reconstruct input. Autoencoder is treated as a generative model and classified as a variational and adversarial autoencoder considering its reconstruction capability.

The ability of autoencoder to handle missing data, its robustness against noisy conditions, and training in an unsupervised manner helps in the implementation of co-learning objectives.

### 5.4 Generative adversarial networks (GAN)

Generative Adversarial Networks [52] have become very popular recently due to their success for unimodal applications such as image synthesis, image to image translation, enhanced image resolution, etc. GAN can generate high-quality samples based on input data distribution in an unsupervised manner. Like other popular methods, GAN is also extended to multimodal applications with the main objective to reduce the distributional difference between the modalities. In cross-modal translation, shared semantics among the modalities is captured, whereas, in cross-modal retrieval, the similarity between modality representation is captured.

In text-to-image synthesis [165], encoded text information is fed along with the noise to a generator to be translated to an image. The discriminator identifies if generated image and encoded text are the same or not. In cross-modal retrieval applications, GANs are used to obtain similar representations in a subspace. The objective is to have a mapping between modalities in a subspace. For instance, the discriminator cannot distinguish the source modality for the features, i.e., similar data from different modalities are next to each other in space. In the training process, the discriminator tries to classify where a feature comes from, and the generator tries to obtain modality invariant representations. Cross-Modal GAN (CM-GAN) [166] creates joint distribution between modalities to promote cross-modal correlation learning for inter-modality and intra-modality correlation for the image-text retrieval, as generators and discriminators act adversely. A cross-modal convolutional autoencoder with weight sharing constraints is used as a generator.

As highlighted in earlier sections of this paper, GAN, along with its variants, is used to generate missing modality [75], handle imbalanced data [106], domain adaptation [121], and noisy conditions [94] for multimodal co-learning.

### *5.5 Multi-task learning*

In Multi-task learning (MTL), models are trained on multiple tasks simultaneously. The use of shared representation to learn multiple related tasks results in data efficiency, reduction in overfitting, and faster computations [54]. However, learning multiple tasks together brings its own difficulties. Each task has different needs, and tuning for one task may impact another task, referred to as *negative transfer*. Designing MTL systems to have positive transfer is an active area of research [167]. In visual systems, MTL focuses on extracting task-specific information and shared information. In language modality, one can ask multiple questions on the same piece of text, enabling MTL. Task-agnostic language representations, like embeddings and pre-trained models being used for MTL.

In multi-task multimodal models, representations are shared across modalities and tasks, thereby increasing generalization. Sentiment and emotions are related tasks that can be modeled using MTL [168] using audio, video, and language modalities. In some instances, high-level tasks are divided into multiple low-level tasks, and models are trained on these tasks using MTL. Scene understanding [169] can be divided into object detection, scene graph generation and region captioning, and joint learning across these three tasks showed better performance over the prevalent models.

For the VQA task [170], a shared encoder with co-attention layers can be developed, and a separate decoder for each task in the MTL set-up. Each task receives representation from its intermediate level, which represents the previous task, thereby creating a *hierarchical* structure. The vision-language tasks considered are image caption retrieval, VQA, and visual grounding with periodic task switching mechanisms during the training phase. A similar concept is recently extended to train 12 tasks together using a multi-task approach [17] using overlapping relationships between different types of vision-and-language (VL) tasks. It uses ViLBERT with a separate head for each task, like branches of a common, shared 'trunk' ViLBERT along with a dynamic stop-go mechanism for each task during training. The use of a large number of datasets created for different individual tasks can be a step towards generalization. Multi-task models can handle missing modalities [6], and multi-task graph convolution can handle audio classification in the presence of noisy labels.

Thus, multi-task learning methods enable multimodal co-learning, supporting objectives of modality availability and noisy conditions.

### *5.6 Transfer learning*

In transfer learning [55], source domain knowledge is transferred to a different but related target domain. Transfer learning helps to reduce demand on labeled data, taking the benefit of an already trained model to improve the performance. However, sometimes there is a negative impact on the learner's ability termed, as a negative transfer. A negative transfer depends on the learner's ability to find transferable and beneficial parts of knowledge and relevance between the source and target domains. Transfer learning is further divided into homogenous and heterogeneous transfer learning based on if domains have the same or different feature spaces [113]. Transfer learning is used for various applications in the vision and language domain by having plenty of pre-trained models and embeddings. Domain adaptation which is one of the topics of transfer learning, is researched now actively.

In the case of *multimodal transfer learning*, it can be a transfer of knowledge from a pre-trained multimodal model or a multimodal embedding to the target domain. It can be a knowledge transfer from one modality to another modality. For example, pre-trained VL models like ViLBERT [155] are trained on large unsupervised data for few tasks, but knowledge can be transferred to multiple other VL tasks with or without fine-tuning. In the teacher-student model, when the teacher is trained on visual modality and the student is trained on sound modality, transfer of knowledge occurs to achieve sound classification using unlabeled video [72]. In DeViSE [83], text information is used to improve visual representation by creating a mapping between visual representation and word2vec [171] embeddings. Heterogeneous Modality Transfer Learning (HMTL) [70] transfers knowledge from text modality, normally a dominant modality in sentiment analysis, to audio and visual modalities.

The embedding layer is trained using adversarial learning so that it is unable to distinguish representations from the source and target model. Transfer learning by creating semantic mappings between embedding layers of audio and visual unimodal networks [172] helps the model to perform audio-video recognition tasks in the absence of audio signals at test time. Knowledge transfer between labeled source domain to unlabeled target domain [173] is used for cross-modal retrieval tasks using pseudo labeling strategy. Similarly, using an auxiliary database with complete modalities enables transfer learning [119] to learn missing modalities in the target using a latent low-rank constraint on subspace.

Thus, transfer learning helps achieve objectives of co-learning by training models on data-rich and clean modalities and transferring that knowledge to data-scarce or noisy modalities. Transfer learning also enables the model to work on missing modalities at test time.

### *5.7 Multimodal embeddings and pre-trained models*

Language embeddings like Word2vec [171] and Gloves [69] are extensively used for various natural language processing tasks. Recently contextual embeddings that assign a representation to each word based on the context have emerged. A family of the pre-trained models like ELMO [174], BERT [51], GPT [175], etc., which are trained on a large corpus, achieved the state-of-the-art performance for language tasks. Likewise, pre-trained models or standard architectures like ImageNet, ResNet, VGG [176], etc., are used in visual domain tasks.

Multimodal pre-trained models like multimodal transformers are becoming popular for multimodal tasks, as discussed in Section 5.2. Similar to language embeddings where similar word representations are considered together, similar representations across the modalities are associated together in the case of multimodal embeddings. Multimodal embeddings are based on cross-modal similarity measurements. Various approaches are used to have a minimum distance between the same semantics samples and a maximum distance between samples with different semantics. In finding similarities, shared representations among the modalities are captured, which can be used to transfer knowledge from one modality to another. Widely used constraints for similarity measurements are cross-modal ranking loss [14] and a Euclidian distance [177]. The visual-semantic embedding model (DeViSE) [83] uses dot product and rank loss to create embedding for visual recognition tasks using text and image modalities. Multimodal embedding showed performance improvement for zero-shot learning. Recently, the use of adversarial learning [166], contrastive learning [178], and meta-learning [56] to obtain better multimodal embeddings. Un-supervised [179] [180], semi-supervised, and weakly supervised [181] learning methods are also used for multimodal embeddings.

Thus, multimodal embeddings and pre-trained multimodal models effectively achieve and bring about the co-learning objectives of missing modality, noisy modality, data parallelism, and modality annotations in the multimodal tasks.

### *5.8 Few-shot learning*

All the data classes are generally seen during training time in machine learning classification models. However, in some situations, there is a need to classify instances whose classes are not seen or very few seen in the training phase. These situations can arise due to a large number of target classes, rare target classes, changing target classes over time, or expensive labeling of target classes [182]. When there is no labeled example at test time, it is called *zero-shot learning* (ZSL) [53], when there is one labeled example at test time, it is called *one-shot* learning [183], and when there are few labeled examples at test time, it is called as *few-shot* learning [184]. Few-shot learning is implemented in three ways [185]: a) augment the *data* with prior knowledge, b) use prior knowledge of *model* to have reduced hypothesis for a function, and c) using prior *algorithmic* knowledge for parameters, regularization, optimization, etc. Few-shot learning and especially zero-shot learning achieved very high performance on visual tasks and are now being investigated for multimodal applications.

For zero-shot image captioning [186], the YOLO [187] model's prediction function is modified to include class embeddings (class names or attributes) to get a zero-shot object detector. Zero-shot object detector and generation template can generate image captions for unseen classes at training time. A new dataset zero-shot transfer VQA [188] is designed so that some words only occur in questions or answers. During testing time, those would be unseen words to VQA model.

Visual representations are projected onto word embedding space to form a joint space [189], which is used to classify objects not seen during the training phase under ZSL settings. Creating a joint embedding space for audio, video, and text labels such that similar classes have lower distances than dissimilar classes [181]. ZSL cross-modal retrieval and classification are achieved using the nearest neighbor search in the embedding space.

Few-shot learning can effectively transfer knowledge from high resource modality to low resource modality and can handle the missing modality at test time. Therefore, few-shot learning enables multimodal co-learning.

### 5.9 Meta-Learning

Meta-Learning is a learning-to-learn algorithm that learns while learning multiple tasks on training data to handle new tasks at test time. Meta-learning improves data efficiency, enables knowledge transfer, and supports unsupervised learning, and it is useful for task-agnostics and task-specific scenarios. Meta-learning is classified into metric-based, model-based, and optimization-based [56]. Transfer learning, multi-task learning, domain adaptation, and meta-learning are often seem similar as all of them try to use prior knowledge. Table 9 highlights the differences among them. Meta-learning became very popular for vision tasks, reinforcement learning tasks, environment learning, neural architecture search, and unsupervised meta-learning. With the success of few-shot learning in language modeling and speech recognition, meta-learning can handle many tasks such as machine translation, low-resource language modeling, etc. Meta-learning is also robust for label noise and adversarial attacks.

**Table 9**: Differences among supervised learning, transfer learning, multi-task learning, domain adaptation and meta-learning methods based on the variation of tasks at training and test time

| Method | Train | Test | Domain Shift |
|---|---|---|---|
| Supervised Learning | Task 1 | Task 2 | No |
| Transfer Learning | Task 1 | Task 2 | No |
| Multi-task learning | Task 1, Task 2, .... Task N | Task 1, Task 2, .... Task N | No |
| Domain Adaptation | Task 1 | Task 1 | Instance-level shift |
| Meta-Learning | Task 1, Task 2.... Task N | Task N+1 | No |

Model-Agnostic Meta-Learning (MAML) algorithm is extended to reconstruct missing modality [59] by using two auxiliary networks to handle severely missing modality at training and testing time. The meta-learning definition is extended to cross-modal generalization [95], having different modalities at source and target. Cross-Modal Meta-Alignment (CROMA) algorithm is used to capture a space where representations of similar concepts in different modalities are close together. Meta-learning is widely used in reinforcement learning, where an agent needs to operate in new target environments. Memory Vision-Voice Indoor Navigation [190] uses voice commands for the agent to move using RGB and depth information with meta-learning.

Thus, meta-learning enables multimodal co-learning by handling missing and noisy modalities.

### 5.10 Multi-instance learning (MIL)

In multi-instance learning (also referred to as multiple instance learning) [57], the dataset consists of labeled bags containing multiple instances. Labels of the instances in the bags are not provided. The bag is labeled as positive if it contains at least one positive instance or negative when all the instances are negative. These assumptions of MIL are relaxed to include a limit on the positive number of instances or a combination of instances [191]. The objective function for MIL is to classify unseen bags accurately and sometimes even classify each instance separately and, in the process, learn a representation that detects positive instances. MIL is used where data is available in sets, and recently, it has become popular for applications with weakly annotated data. For audio-video classification, MIL utilizes weak supervision among them. MIL algorithms are classified into three types – instance space algorithms, bag-space algorithms, and embedding-space algorithms. Based on the success of MIL for unimodal tasks, it is now becoming popular for multimodal applications, especially where weakly annotated data is available. The use of weakly annotated data reduces the labeling cost, and large data becomes available mostly from the web.

Large-scale sound classification is performed using weak supervision between audio-video with MIL [192]. Each video segment is treated as a bag, and frames in a video are treated as instances; similarly, for audio modality, each segment is treated as a bag and frames with Mel spectrograms as instances. There is no labeling at the frame's level, but video segments are labeled with the sound it contains.

Complex objects [193] like an image with multiple objects can be divided into multiple regions with Fast-RCNN, and corresponding tags can be divided into multiple word vectors. Regions and word vectors become instances of bags of image and caption. Weak supervision between image and caption, i.e., bag level, is used with multi-instance learning for object classification. Detection of violent videos [194] by classifying videos based on video scenes and audio sounds into positive and negative bags and using associated text attributes in the set-up of multimodal MIL is developed.

Thus, multi-instance learning, which supports weakly supervised learning without explicit data alignment, is helpful to achieve co-learning objectives.

### *5.11 Contrastive learning*

In *self-supervised learning*, an internal representation of the data is created for the downstream tasks. For example, more images and labels can be prepared by image augmentation like rotations, cropping, contrasting, and colorization. These augmented images form a pair of similar images and are dissimilar from the rest of the images in a dataset. If we can learn a representation function on this data, it can help in the downstream task. These representations are learned by *contrastive learning* using contrastive triplet loss and sub-sampling of negative samples. A feature memory bank [195] to store the representation of images instead of computing every time and *Momentum Contrast* (MoCo) [196] with a dictionary as a queue of data samples are the latest sub-sampling techniques.

*SimCLR* [197] proposed a simple contrastive framework for visual representations that do not need a memory bank or special architectures. It is achieved by choosing effective data augmentation of images, namely, random cropping-resizing back to the original size, random color distortion, and random Gaussian blur. Two encoders are trained to maximize contrastive loss for a similar instance, and in the process, the best representation is obtained, which is used for downstream tasks. Contrastive learning is extended to patch level within an image to achieve unpaired image-image translation [198]. Adversarial perturbation of images creates more samples for contrastive learning.

The success of contrastive learning in unimodal tasks and its principle of using multiple views of single input make it suitable for multimodal data. Using two versions of MNIST [199] and brain imaging dataset [18] proved that multimodal contrastive learning performs better than unimodal with a proper choice of the loss function and adds a regularization effect. The contrastive learning applied to multi-view settings [200] to maximize the representation of two views outperforms popular multiview methods.

For multimodal VL applications, models create embeddings for each modality in a common space. However, these embeddings cannot address inter-class dynamics and intra-class relationships. Contrastive Bimodal Representation Algorithm (COBRA) uses contrastive learning to create joint cross-modal embeddings [201]. The projected representation of data samples belonging to the same class and the same modality is considered positive pairs. The projected representations of data samples belonging to different classes of the same or different modalities are considered negative pairs. Noise Contrastive Estimation is used in a loss function.

Multimodal generative models make use of the commonality shared between the related pairs; however, the information from the unrelated pairs information is not utilized; contrastive learning uses both related and unrelated pairs information [202], thereby reducing the need for paired multimodal data to achieve the same level of performance as that of a generative model.

Vision language pre-training models like ViLBERT [155] and LXMERT [153] use feature regression or classification based on visual region and masked word. These models inherit the bias and suffer from noisy labels of the pre-training dataset. It is overcome by Contrastive Vision-Language Pretraining (CVLP) model [178] using contrastive loss on the visual branch. CVLP outperformed ViLBERT, LXMERT, and other pre-trained vision-language models on VQA datasets.

In phrase grounding, the association between image and region and a phrase and caption is required, preferably with weak supervision. Creating negative samples by substituting words in captions [108] and then maximizing the mapping between regions and corresponding captions against non-related regions with contrastive learning and captions showed better performance for weakly phrase grounding. The contrastive learning to align the representation of modalities by maximizing the agreement among them for music genre classification [203] improves the model's accuracy over regular prediction-based models or collaborative filtering.

Thus, multimodal co-learning is benefiting from contrastive learning for performance and weakly supervised data from the web.

### *5.12 Domain adaptation*

As discussed in Section 4.5, applications like emotion recognition and image and cross-modal retrieval [116], social media event rumor detection [117] using disentanglement, multimodal sentiment analysis with text as a source and audio-video as target domain [70], and audio-visual cross-modal mapping [121] using dacssGAN (Domain Adaptation Conditional Semi-Supervised Generative Adversarial Network) used domain adaptation.

The complete modalities auxiliary database [119] and its cross-modality and cross-database knowledge addresses missing modalities. The auxiliary data could be co-occurrence data or embedding spaces formed by projecting source and target domain data samples [113]. Domain translator function [120] learned from multimodal source data can predict class even if one modality is absent using weakly shared deep transfer networks. Cross-modal generalization [95], e.g., the source is an image and the target is audio, is achieved using cross-modal meta-alignment with contrastive learning.

Thus, homogenous and heterogeneous multimodal domain adaptation and generalization methods support missing modalities, noisy conditions, lack of annotated data and use of cross-domain knowledge.

### *5.13 Multimodal co-learning computational methods summary*

In this section, we discussed various methods for multimodal co-learning implementations. Multimodal deep learning is adapting all the latest deep learning methods and models. The fusion, attention, and transformers-based methods provide a better representation of multimodal data and achieve higher accuracies addressing the challenge of having strongly paired data. Attention networks enable working with weakly supervised data. Graph-based networks use domain knowledge to work with hybrid data relationships and weak supervision. Encoder-decoder, autoencoder, and multi-task learning can handle missing modalities at test time. The multimodal embedding helps in non-parallel data, semi-supervised, and unsupervised data conditions. Disentanglement and counterfactuals enable interpretability and fairness in models. Transfer learning and meta-learning for domain adaptation, adversarial learning for building robustness against noise, and contrastive learning for unsupervised data models. GAN-based models to support missing modality, noise robustness, and domain adaptation. The summary of methods is shown in Table 10.

**Table 10**: Summary of multimodal co-learning computational methods

| Sr. No. | Implementation Methods | Multimodal Taxonomy Challenges | | | | | Presence of Modality | | | | Data Parallelism | | | Noisy Modality | | Modality Annotations | | | | Domain adaptation | | Interpretability & Fairness | | References |
|---|---|---|---|---|---|---|---|---|---|---|---|---|---|---|---|---|---|---|---|---|---|---|---|---|
| | | Representation | Alignment | Translation | Fusion | Co-learning | Fully Missing | | Partly Missing | | Parallel | Non-Parallel | Hybrid | Data Noise | Label Noise | Supervised | Semi-supervised | Un-supervised | Weakly supervised | Domain Adaptation | Domain General. | Inter. & Explain. | Fairness & Bias | |
| | | | | | | | Train | Test | Train | Test | | | | | | | | | | | | | | |
| 1 | Multiple Kernel Learning | ■ | | | | ■ | | ■ | | | ■ | | | | ■ | | ■ | | ■ | | | | | [48] |
| 2 | Co-training | ■ | | | | ■ | | ■ | | | ■ | | | | | | | | ■ | | | | | [47] |
| 3 | Teacher-Student | ■ | | | | ■ | | ■ | | | | ■ | | | | | | | ■ | | | | | [72] |
| 4 | Prototype Network | ■ | | | | ■ | | ■ | | | ■ | | | | | ■ | | | | | | | | [9] |
| 5 | Knowledge Distillation & Privileged Learning | ■ | | | | ■ | | ■ | | | ■ | | | ■ | | ■ | | | | | | | | [65] |
| 6 | Bridge Correlational Neural Networks | ■ | ■ | | | ■ | | ■ | | | | | ■ | | | ■ | | | | | | | | [10] |
| 7 | Multimodal Factorization | ■ | | | ■ | ■ | | ■ | | | ■ | | | | | ■ | | | | | | | | [67] |
| 8 | Multimodal Embedding | ■ | ■ | | | ■ | | ■ | | | ■ | | | | | ■ | | | | | | | | [68] |
| 9 | Multimodal Embedding | ■ | | | | ■ | | ■ | | | | ■ | | | | | ■ | | | | | | | [83] |
| 10 | Multimodal Embedding | ■ | | | ■ | ■ | | ■ | | | | | ■ | | | ■ | | | | ■ | | | | [87] |
| 11 | Multi-task Learning | ■ | | | | ■ | | ■ | | | ■ | | | | | ■ | | | | ■ | | | | [17] |
| 12 | Multi-task Learning | ■ | | | ■ | ■ | ■ | ■ | | | ■ | | | | | ■ | | | | | | | | [6] |
| 13 | Memory Fusion Network | ■ | | | ■ | ■ | | ■ | | | ■ | | | | | ■ | | | | ■ | | | | [15] |
| 14 | Fusion | ■ | | | ■ | | | | | | ■ | | | | | ■ | | | | | | | ■ | [141] |
| 15 | Fusion – Pre-training | ■ | | | ■ | | | | | | ■ | | | ■ | | ■ | | | | | ■ | | | [122] |
| 16 | Autoencoder | ■ | | | ■ | ■ | | ■ | | | ■ | | | | | ■ | | | | | | | | [24] |
| 17 | Autoencoder | ■ | | | | | | | | | ■ | | | | | ■ | | | | ■ | | | | [123] |
| 18 | Encoder-Decoder | ■ | ■ | ■ | | ■ | | ■ | | | ■ | | | | | ■ | | | | | | | | [23] |
| 19 | Encoder-Decoder | ■ | | | ■ | ■ | | ■ | ■ | | ■ | | | ■ | | ■ | | | | | | | | [77] |
| 20 | Encoder-Decoder | ■ | ■ | ■ | | ■ | | ■ | | | ■ | | | | | ■ | | | | | | | | [21] |
| 21 | Encoder-Decoder | ■ | | ■ | | ■ | | ■ | | | | | ■ | | | ■ | | | | | | | | [85] |
| 22 | Encoder-Decoder | ■ | | | ■ | ■ | | ■ | | | ■ | | | ■ | | ■ | | | | | | ■ | | [133] |
| 23 | Multi-Instance Learning | ■ | | | | ■ | | ■ | | | | ■ | | | ■ | | | | ■ | | | | | [79] |
| 24 | Graph-based Network | ■ | | | ■ | ■ | | ■ | | | | | ■ | | ■ | | ■ | | | | | | | [91] |
| 25 | Graph-based Network | ■ | ■ | | | | | | | | | | ■ | | | | | ■ | | | | | | [86] |
| 26 | Transformers | ■ | ■ | | | ■ | | ■ | | | | | ■ | | | | | | ■ | | | | | [84] |
| 27 | Transformers | ■ | ■ | | ■ | ■ | | ■ | | | | ■ | | ■ | | | | | ■ | | | | | [111] |
| 28 | Attention Network | ■ | | | ■ | ■ | | | | | ■ | | | ■ | | | | | ■ | | | | | [102] |
| 29 | Meta-Learning | ■ | | | | ■ | | ■ | ■ | ■ | ■ | | | | | | | | | | | | | [59] |
| 30 | Meta-alignment - NCE | ■ | ■ | | | ■ | | ■ | | | ■ | ■ | | | ■ | ■ | | | | | ■ | | ■ | [95] |
| 31 | Adversarial Learning | ■ | | | | | | | | | ■ | | | ■ | | ■ | | | | | | | | [103] |
| 32 | Adversarial Learning | ■ | | | ■ | | | | | | ■ | | | | | ■ | | | ■ | | | | | [118] |
| 33 | Contrastive Learning | ■ | | | | ■ | | ■ | | | | ■ | | | | | | ■ | | | | | ■ | [142] |
| 34 | Contrastive Learning | ■ | ■ | | | | | | | | | ■ | | | | | | ■ | ■ | | | | | [22] |
| 35 | Contrastive Learning | ■ | ■ | | | ■ | | ■ | | | | ■ | | | | | | | ■ | | | | | [108] |
| 36 | Transfer Learning | ■ | ■ | | ■ | | | | | | ■ | | | | | | ■ | ■ | | ■ | | | | [109] |
| 37 | Transfer Learning | ■ | ■ | | | | | ■ | | | ■ | | | | | ■ | | | | ■ | | | | [119] |
| 38 | Transfer Learning | ■ | ■ | | ■ | ■ | | ■ | | | | ■ | | | | ■ | | | | ■ | | | | [70] |
| 39 | Transfer Learning | ■ | ■ | | | ■ | | ■ | | | | | ■ | | | | | | ■ | | ■ | | | [126] |
| 40 | GAN | ■ | ■ | | ■ | ■ | | ■ | | | ■ | | | | | ■ | | | | ■ | | | | [116] |
| 41 | GAN | ■ | | | | ■ | | ■ | | | ■ | | | | | ■ | ■ | | | ■ | | | | [121] |
| 42 | GAN | ■ | | | ■ | | | | | | ■ | | | | | ■ | | | | | | | | [106] |
| 43 | GAN | ■ | | | | | | | | | ■ | | | | ■ | ■ | | | | | | | | [94] |
| 44 | GAN | ■ | | | | ■ | | ■ | | | ■ | | | ■ | | ■ | | | | | | | | [75] |
| 45 | GAN | ■ | | | ■ | ■ | ■ | ■ | | | ■ | | | | | | ■ | | | | | | | [11] |
| 46 | GAN | ■ | ■ | | | ■ | | ■ | | | ■ | | | | | | | | ■ | | | | | [124] |
| 47 | Disentanglement | ■ | | | | | | | | | ■ | | | | | ■ | | | ■ | ■ | | | | [117] |
| 48 | Disentanglement | ■ | | | | | | | | | | | | | | | | ■ | | | | ■ | | [134] |
| 49 | Counterfactual | ■ | | | | | | | | | ■ | | | | | | ■ | | | | | ■ | | [136] |
| 50 | Counterfactual | ■ | | | ■ | | | | | | | ■ | | | | | | | ■ | | | ■ | | [137] |

## 6. Datasets and Applications of multimodal co-learning

Applications from various domains use multiple modalities to improve the performance of tasks involved. These applications and tasks must support multimodal co-learning objectives to cater to real-life situations. Multimodal co-learning is used for applications ranging from classification, regression, action recognition, event detection, sentiment and emotion detection, speech enhancement, sound classification, remote sensing, rumor detection, deception detection, recruitment, vision-language tasks, VQA, phrase grounding, machine translation, cross-modal retrieval, etc.

Here in Table 11, we summarize representative applications from our investigation of research studies. The details of applications with their categories, modalities involved, methods used, and the corresponding datasets are presented.

**Table 11**: Summary of multimodal co-learning applications and datasets

| Task Categorization | Application/ Task | Modalities Involved | Methods | Datasets | Notes | References |
|---|---|---|---|---|---|---|
| Classification | Image Classification | Image, Text | Multiple Kernel Learning | PASCAL VOC'07, MIR Flickr | Semi-supervised image-tags | [48] |
| | | Image | Meta-alignment & NCE | CIFAR-10, CIFAR-100 | Robust to label noise | [94] |
| | | Image | Disentanglement | MNIST, SVHN, CelebA | Maximizing mutual information - InfoGAN | [134] |
| | | Image, Text | Multimodal Embedding | ImageNet (ILSVRC) 2012 1K | Image labels and un-annotated tags | [83] |
| | Object Classification, Action recognition | Image, Text | Multimodal Embedding | 400 Million Image-Text pairs | Cosine similarity | [142] |
| | Video Classification | Audio, Video, Text | Co-training | Fudan-Columbia Video Dataset (FCVID) | Video meta-data as textual-modality | [47] |
| | Sound Classification using Video | Audio, Video | Teacher-Student | 2M Videos from Flickr & YouTube | Object and scene detectors used to extract from videos | [72] |
| | Classification of web pages | Text, Image, HTML layout | GAN | Faculty Web pages | Imbalanced data handling | [106] |
| | Audio tagging | Audio, Ontology | Graph-based Network | Google Audio Dataset | Ontological relationship between sound events | [91] |
| | Remote Sensing | SAR Image, Optical Image | Prototype Network | Multi-Sensor All-Weather Mapping (MSAW) dataset | Meta-sensor representation | [9] |
| | Video activity Recognition | RGB, Depth | Knowledge Distillation & Privileged Learning | NTU RGB+D | Hallucination Network to generate missing modality data | [65] |
| | Audio Visual Event Recognition | Audio, Video | Attention Network | Google Audio Dataset | Adversarial noise is added to audio modality | [102] |
| | Sentiment Analysis | Audio, Video | Autoencoder (AE) | MOSI & MOSEI | Reconstruct missing modality using AE | [24] |
| | Sentiment Analysis | Audio, Video, Text | Transfer Learning | MOSI, IEMOCAP | Text as source modality for domain adaptation | [70] |
| | Sentiment Analysis | Audio, Video, Text | Multimodal Factorization | CMU MOSI, POM, MNIST, SVHN | Discriminative and generative factors | [67] |
| | Sentiment Analysis | Audio, Video | Encoder-Decoder | RECOLA | Translation based representation | [21] |
| | Sentiment Analysis | Audio, Video, Text | Memory Fusion Network | CMU MOSI, POM, IMDB, SST | Fusion and Attention on LSTM | [15] |
| | Sentiment Analysis | Audio, Video, Text | Encoder-Decoder | MOSI, ICT-MMO, YouTube | Cyclic Translation loss | [23] |
| | Sentiment Analysis | Image-Text Audio-Image | Meta-learning | Multimodal IMDB, MOSI, Audiovision MNIST | Modality complete dataset is required | [59] |
| | Emotion Prediction | Image, Text | Multi-task Learning | Flickr Emotion, Visual Sentiment Ontology (VSO) | Multi-task like model and fusion | [6] |
| | Emotion Prediction | Audio, Video | Fusion and Adversarial learning | MSP-IMPROV | Acted and improvised emotions are considered a domain shift. | [118] |
| | Emotion Prediction | Audio, Video | GAN | CREMA-D, RAVDEES | Conditional GAN | [121] |
| | Emotion Prediction | Audio, Video | Pre-training and Fusion | EmotiW, ChaLearn | Pre-training on VGG Face and finetuning on FER | [122] |
| | Emotion Prediction, Cross-media Retrieval | a) Audio, Video b) Image, Text | Domain Adaptation | IEMOCAP & AFEW, MSCOCO & CUB-200 | Attention network | [116] |
| | Gait Recognition | Motion, RGB-D, EMG | Multimodal Autoencoder | Abnormal Gait Dataset | Cross-subject, cross-modal transfer learning for domain generalization | [123] |
| | Deception Detection | Audio, Video | Transfer learning | Videos of a real court trial for high stake and UR Lying as low stake deception | Domain adaptation from low stake to high stake deceptions | [109] |
| | Face Recognition, Image Classification | Image, different domain Image | Transfer Learning | CMU-PIE and Yale B Face, BUAA and Oulu VIS-NIR Face, ALOI-100 and COIL-100 Dataset | Heterogeneous domain adaptation | [119] |
| | Speech Enhancement | Audio, Video | Encoder-Decoder | Video of a native speaker for 320 Mandarin sentences | Fusion | [77] |
| | Modality generalization | Image, Audio | Encoder-Decoder | Text-Image: Yummly-28K Image-Audio: CIFAR & ESC50 Text-speech: Wilderness | Meta-alignment algorithm with Noise Contrastive Estimation. | [95] |
| Generation | VL Tasks | Image, Text | Multimodal Embedding | Visual Genome, CIFAR 100 | Visual co-occurrences between object and attributes words. | [68] |
| | VL Tasks | Image, Text | Multi-task Learning | 12 datasets from VQA tasks | ViLBERT as an underlying model. | [17] |
| | VL Tasks | Image, Text | Adversarial Learning | 9 VL Datasets | Noise injecting in embedding space. | [103] |
| | Image-sound generation | Image, Audio | Cycle GAN | Sub-URMP (University of Rochester Musical Performance) dataset | Four encoders-decoders for generating all combinations | [75] |
| Translation | Language Translation | Image, Text | Bridge Correlational Networks | Multilingual TED corpus, MSCOCO | Image is a pivot modality, and source-target are texts. | [10] |
| | VQA | Image, Text | Counterfactual | VQAv1 | Counterfactual images are generated. | [136] |
| | Machine Translation, Crosslingual Image description | Image, Text | Graph-based Network | Multi30K extension of Flickr30K for English and German language | Source language as pivot modality for an image description. | [85] |
| Alignment | Alignment Task | Video, Text | Graph based Network | Microsoft Research Multimodal Aligned Recipe Corpus | Alignment is created for text-text, video-video, and text-video using text and video transcript. | [86] |
| | Multimodal Embedding | Audio, Image, Text | Autoencoder | Glove, ImageNet with WordNet synsets, Freesound for audio, Word association dataset | Word association is used to create alignment among datasets | [87] |
| Cross-Media Retrievals | EEG-Image retrieval | EEG, Image | Cycle GAN, Triangle GAN | ImageNet-EEG | Shared latent space is divided into semantic and semantic-free latent variables. | [11] |
| | Text-Video retrieval | Video, Text | Multi-Instance Learning | HowTo100M | ASR is used for alignment. | [79] |
| | Multilingual Video Search | Video, Text (9 languages) | Transformers | Multi-HowTo100M (1.2 M videos with nine language subtitles | Visual concepts are used as a pivot for all language sentences. | [84] |
| | Text-Video retrieval | Audio, Video, Text | Multimodal Transformers with MIL-NCE | YouCook2, MSRVTT | Weights are shared among three modalities with separate tokenization and projection heads, | [111] |
| | Image-Text retrieval | Image, Text | Cross modal data programming with a generative model | Medical dataset with image and text | Rule-based weak supervision using an auxiliary dataset | [124] |
| | cross-modal (event) retrieval | Image, Text | Transfer learning | MMED | Cross-domain cross-modality transfer learning with dictionary alignment | [126] |
| | Cross-Media Retrieval | Image, Text | Encoder-Decoder | 0.72 million fashion products and accessories | Disentanglement based on fashion attributes such as color, style, etc. | [133] |
| Phrase Grounding | Phrase Grounding | Text, Image | Contrastive Learning | Flickr30k | Pre-trained object detectors to extract objects. | [22] |
| | Phrase Grounding | Text, Image | Contrastive Learning | Flickr30k, MSCOCO | Contrastive learning by adding negative words in captions | [108] |
| | Phrase Grounding | Image, Text | Counterfactual | COCO2017 dataset and Flickr30k Entities dataset | Decomposition helps in interpretability and Counterfactuals in robustness | [137] |
| Social Media | Rumor Detection | Image, Text | Disentanglement | PHEME and PHEME_veracity | Disentanglement into content and rumor style | [117] |
| Human Resource | Recruitment | Image, Text | Fusion | 24,000 resumes with faces, text | Gender Ethnicity Bias | [141] |

## 7. Open problems and future research directions

Tackling real-life conditions is of paramount importance for any computational application. Unimodal deep learning systems can meet some of these requirements. However, for multimodal applications, the capability to work in real-life situations is emerging in which multimodal co-learning plays a vital role. While there are some advancements in tools and techniques for implementing multimodal co-learning objectives, there are many exciting research problems that need to be solved. Multimodal co-learning is constrained by the current multimodal problems like universal representation, multimodal dataset availability, and evaluation tools. The solution to co-learning open problems will also help to resolve some of the issues encountered for multimodal architecture development and implementation. In this section, we present open problems and future directions.

### *7.1 Multimodal representation for co-learning*

Vision, language, audio, video encoders, pre-trained models, or word embeddings are used for representation. Multimodal deep learning systems capture inter-modality information which is complementary or supplementary to each other. However, there is a challenge to have a universal representation of multiple modalities that can take care of complementary and supplementary information preserving intra-modality structures. At present, two prevalent techniques - joint representation and coordinated representation cannot meet this objective in spite of enhancing those with the latest deep learning techniques. Most of the multimodal representations proposed are designed for domain-specific data inputs and are not effective on different data combinations.

There is also a need for generic representation for co-learning, which is easy to implement and provides better performance to missing and noisy modalities scenarios. The representation should be configurable to support different types and a number of modalities for co-learning.

Creating *multimodal embeddings* and enhancing them using the latest deep learning techniques is one approach to a have better representation of multiple modalities in a joint space. The issue of preserving intra-modality similarity or dissimilarity structure is addressed using a discriminative algorithm. *Generative adversarial network (GAN),* which consists of generator-discriminator mechanism, offers solutions as generators try to generate modality invariant and a discriminator distinguishes the features from different modalities. Thus, multimodal embeddings with pre-trained models and GAN frameworks can be focused on arriving at generic representations which can support multiple and different types of modalities.

### *7.2 Multimodal dataset availability*

Image and language modalities have achieved higher performances due to the progress in deep learning and the availability of large and quality datasets. Pre-trained models are developed to transfer knowledge for new applications with fewer data and achieve required performances. However, pre-trained models are not yet available for audio and video modality as there is a lack of large audio-video datasets. It has led to either audio-video models getting overfitted or relying on language modality as a dominant modality.

The available multimodal datasets are mostly generated and collected in a controlled setting and do not truthfully represent real-life situations truthfully [204]. For example - videos for emotion detection are in recorded controlled conditions, videos available on YouTube for activities like cooking recipes, reviews, conversations, debates are all recorded in certain controlled conditions. Current studies drop samples or modalities from the well-processed datasets to simulate missing modality scenarios. There is a need for a real-life dataset of missing modalities that may contain noise or low power signals. The different training strategies will require to implement co-learning on real-life datasets.

Multimodal datasets are available for visual-language tasks, sentiment and emotion classification, instructional videos, social media, etc.

There is a lack of multimodal datasets in other fields like product recommendation engines, mechanical machine conditions monitoring, agricultural field and crop conditions, medical conditions, transportation, chemical engineering, communication, etc. All these fields have a multimodal nature of data; in fact, it is more complex than the currently available multimodal datasets. These domains are more prone to sensor faults, dust, humidity, weather conditions, etc., which adds noise and impacts the availability of modalities. The multimodal co-learning will help to achieve robustness and avoid industrial accidents causing severe economic losses, environmental impact, and harm to human lives. Hence, the datasets with modalities – EEG, ECG, gas sensors, chemical sensors, network traffic, thermal images, industrial machine health, satellite, crop yield prediction, crop diseases, etc. are required to expand co-learning in these domains.

The availability of multiple datasets for similar tasks [17] enables models to achieve generalizability. Hence, there is an urgent need to have large multimodal datasets in different areas. However, it is challenging to have a multimodal dataset as each modality with a different structure and content often demands particular expertise for annotation. The initiatives using unsupervised, semi-supervised, and weakly supervised techniques are in progress and should be increased to obtain large-scale datasets using free and abundant data available on the internet [86]. Pre-trained multimodal models should be used CLIP [142] to obtain large paired data from the internet without manually aligning and labeling.

### *7.3 Evaluation and diagnostic tools*

Multiple evaluations and diagnostic tools and techniques evolved with the continuous improvements in deep learning models and algorithms. For image and language modalities, evaluation methods and diagnostic tools are standardized now. Areas like interpretability, fairness, and bias also started designing and developing evaluation and diagnostics tools. Likewise, increased research efforts are required in designing and developing evaluation and diagnostic tools for multimodal models. The tools to measure the contribution of each modality in the final prediction will also help evaluate co-learning.

At present, the most widely used evaluation criterion is the comparison of multimodal model results with unimodal baselines, which is an important measurement but not sufficient to understand the contribution of multimodality. The performance metrics such as accuracy, precision, recall, F1-scores, which were designed for unimodal, are also used for multimodal and same are used for co-learning performance measurement. Recently, Empirical Multimodally-Additive function Projection (EMAP) [205] diagnostic tool is proposed to analyze multimodal classification models. In this tool, less expressive space is used to project the multimodal classifier's prediction to be compared with an ensemble of individual unimodal classifiers. It is observed that even the performance gain of models like transformers with cross-modal attention is not fully due to multimodality. It is recommended that all multimodal classifier models should report EMAP along with multimodal and unimodal classifier baselines.

In the multimodal machine translation model [206], to check if the visual modality is beneficial, language input and visual input are gradually degraded. It is observed that visual modality is useful only when language modality is noisy. It is one step towards understanding when to use additional modalities for multimodal applications. Similar evaluation will also help to test the co-learning objective of noisy modality. Taking the clue from how probes are intended for language embedding, multimodal embedding probes [207] are designed to know if visual information is used by language modality.

The efforts towards developing succinct evaluation techniques and appropriate diagnostics tools are essential. These tools should evaluate the effectiveness of co-learning techniques for each of its objectives. The co-learning evaluation approach should ascertain the number and combination of modalities during training, acceptable amount and types of noise, amount of pairing required, type of domain shifts, and interpretability and fairness index.

### *7.4 Presence of modality*

The primary objective of multimodal co-learning is to perform even if one or more modalities are missing at test time against the number of modalities used during training. It is a very important aspect of multimodal applications as in real-life conditions, often all modalities are not present test time, i.e., at actual usage. Based on our investigation, very few multimodal models are designed to meet this objective, although this has been mentioned as one of the challenges in early studies of multimodal deep learning.

Lately, the research community has started to focus on the objective of missing modality so that models can perform within an acceptable level without any adverse impact on accuracy and robustness. The absence of modality at test time is addressed in some of the applications by the nature of the application itself. E.g., VL applications such as image captioning and cross-media retrieval have one modality at test time. Training models for tackling missing modality is also helpful to obtain modality-invariant representations. Interpretability, fairness, and robustness of models are increased if designed for managing missing modalities. The open problems and future directions for the presence of modality co-learning objectives are presented in Table 12.

**Table 12**: Open problems and future directions for the presence of modality objective

| Problem Area | Problem Description | Future Directions |
|---|---|---|
| Modality Availability at training or testing time | • Few studies are designed to handle missing modality that too at test time only.<br>• No recommended guidelines on how to handle missing modalities is available | • Design and test the multimodal models to handle missing modalities at test time.<br>• Increase focus on missing modality at training time with GAN, meta-learning, and domain adaptation techniques.<br>• Compare the methods to handle missing modalities and create a baseline. |
| Partly missing modalities | • Models are not designed for a real-world scenario of alternatively or partly present modalities, as shown in Fig. 6g and 6h. | • Design special training mechanisms like multi-style or dynamic stop-and-go train models on expected and random scenarios. |
| Dominant modality | • Lack of guidance on which modality and when it is supportive at train and test time.<br>• Many studies rely on already-dominant text modality for co-learning | • Models should be evaluated for SEW and WES combinations to decide on modalities to be used.<br>• Modality conditions must be varied to know when it is actually supportive. |
| Auxiliary Datasets | • Auxiliary modality-complete datasets are not available, which are required as a prior to handle missing modality using meta-learning or domain adaptation techniques. | • The relationships must be established among the application-wise existing datasets to act as auxiliary datasets. It is available for some vision-language datasets. |

### *7.5 Data parallelism*

Current multimodal deep learning models demand alignment among the constituent's unimodal signals, and often alignment is created manually. Audio, video and text modalities datasets are manually transcribed to extract spoken words and each utterance's start time and end time. The text is aligned to audio at phoneme and word-level using a forced aligner like P2FA. This data pre-processing step creates a restriction on using these multimodal models in real-time and takes significant effort.

Attention models can learn the relationship between non-aligned multiple modality data. It helps to increase the performance of models as well as the need to have fine-grained alignment. Attention is also used to understand the intra-modality and inter-modality relationship. Attention coupled with memory networks can handle long-duration temporal signals as they can remember important clues for predicting that sequence. However, the success of attention is limited to two modalities data, and it is challenging to handle three and more modalities signals.

Finding similarities and constructing a multimodal embedding using coordinated representation is also perceived as an alternative to strong pairing conditions. Lately, contrastive learning has been used to achieve better alignment as contrastive learning uses negative (unpaired) samples along with positive ones from the dataset.

Another approach to address alignment is to use higher-level alignment, which is either easy to create or available intrinsically in the data. For example, a textual recipe and a video of the recipe are aligned at a higher level as both will have the same or similar steps even though those are not exactly aligned. This relationship is used by techniques like MIL and attention avoiding explicit finer alignment. Another example is phrase grounding, where alignment at image and caption creates a mapping of objects in image and phrases in a caption. Hybrid data or shared data can be treated as a third modality or a view of modality, which aligns with the other two modalities even though those two modalities are not aligned. The open problems and future directions for the data parallelism objective are listed in Table 13.

**Table 13**: Open problems and future directions in data parallelism objective

| Problem Area | Problem Description | Future Directions |
|---|---|---|
| Parallel or strongly paired data | • Offline pre-processing and alignment at a finer level, i.e., word level, utterance level, is required, which is time-consuming, error-prone as well as restricts real-time online usage. | • Extend attention and memory networks to more than two modalities.<br>• Multimodal embedding and coordinated representations along with contrastive learning should be explored. |
| Non-parallel or weakly paired data | • No established methods to create models using weakly paired modality data.<br>• In many cases, weak relationships are not present among the modalities. | • Use pre-trained models like object detectors to create pairing.<br>• Graph networks that capture domain relationships or MIL are promising and are to be explored. |
| Hybrid or shared data | • There is limited availability of datasets that have shared relationships. It is available mainly for multilingual translation tasks. | • Create shared relationships using pivots, e.g., visual modality is a pivot for captions in multiple languages to have alignment and supervised relationships.<br>• Use task-specific relationships, e.g., sentiment and emotions can be related or multi-tasks applications with primary and secondary tasks. |

### 7.6 *Noisy modality*

Noisy labels impact the robustness and performance of the machine learning models, and the methods to address those are evolving. As discussed in Section 3.1, the co-teaching family of networks handles noisy labels by training on clean labels first and using that network to classify remaining samples. Generative models, graphical models, adversarial and contrastive learning are being used to handle label noise and data noise conditions. We have observed that models are not designed to be robust against various noisy conditions, and there are no uniform practices across applications and modalities. We outlined open problems and future directions in Table 14.

**Table 14**: Open problems and future directions in noisy modality objective

| Problem Area | Problem Description | Future Directions |
|---|---|---|
| Label Noise | • Methods to handle label noise are not established for multimodal models.<br>• Lack of noisy labeled multimodal datasets like unimodal ones. | • Develop multimodal models for different label noise conditions.<br>• Use contrastive learning, adversarial training, meta-learning, MIL methods to handle noisy label conditions.<br>• Design proper loss functions to handle label noise. |
| Data Noise | • Simulating noisy conditions and noise levels require a domain | • Develop application-specific noise generator APIs[4] as a part of deep learning libraries. |
| | expertise and differs from task to task. | • Using these APIs, train and test the models to arrive at a trade-off between accuracy and robustness to noise.<br>• Measure and report acceptable noise types and their limits for specified tasks and models.<br>• Use GAN, contrastive learning, adversarial training, and modality-specific noises to generate noisy data.<br>• Create noisy multimodal datasets similar to ImageNet-C, Tiny ImageNet-C, and CIFAR10-C. |

[4] *APIs – Application Programming Interfaces*

### *7.7 Modality annotations*

Availability of annotated data in large quantities is one of the driving factors in the success of deep learning models. There are good datasets available for image and language modalities. However, the available multimodal datasets are still limited to some common tasks, and there is a need to have datasets for applications in other domains. But it is not easy to annotate data; hence there is ongoing research to benefit from semi-supervised, weakly supervised, and unsupervised learning. We highlight some of the problems in this area and future directions to address those in Table 15.

**Table 15**: Open problems and future directions for modality annotations objective

| Problem Area | Problem Description | Future Directions |
|---|---|---|
| Semi-supervised | • Often it is assumed that the unlabeled part has the same class distribution as the labeled part of data.<br>• Need approach when one modality is labeled and the other is unlabeled | • Develop multimodal semi-supervised learning methods to handle class imbalance and bias and test those for missing and noisy modalities.<br>• Measure model performance with a single modality, both labeled, and one labeled and another unlabeled. This is to check if domain knowledge in unlabeled modality supports co-learning. |
| Weakly supervised | • It needs to identify relationships among the modalities, which may not be possible or correct.<br>• It uses standard tools (like pre-trained object detectors to capture objects in an image), and overall performance depends on standard tools' performance. | • Experiment with different degrees of supervision to arrive at the preferred level (i.e., component, concept).<br>• Create recommendations on the type of knowledge (e.g., objects, color, context) to be extracted from available modalities to create supervised relationships. |
| Un-supervised | • The performance of the unsupervised multimodal models depends upon the kind of supervised information available in the data structure itself. | • Enhance established techniques from vision domain to multimodal domain, i.e., contrastive learning with data augmentation without any supervised data. |

### *7.8 Multimodal domain adaptation*

Domain adaptation is a step towards the generalization of machine learning models that can use available data from one domain to predict tasks with fewer data from another domain. Multimodal domain adaptation can take advantage of a large amount of multimodal data available on the web. Multimodal domain adaptation fulfills co-learning objectives as it deals with missing and noisy modalities supporting different distributions in source and target domains. The open problems and future directions in multimodal domain adaptation are as in Table 16.

**Table 16**: Open problems and future directions in domain adaptation objective

| Problem Area | Problem Description | Future Directions |
|---|---|---|
| Domain Datasets | • Availability of multiple domains labeled datasets is a challenge for domain adaptation's success.<br>• Domain datasets with different modalities at source and target are not available. | • Create multiple domain multimodal datasets using weakly supervised and unsupervised multimodal deep learning.<br>• Curate auxiliary datasets with relationship to source and target datasets. These are required as a prior for meta-learning. |

### *7.9 Multimodal interpretability and fairness*

Users demand reasons for machine learning-based decisions across walks of life. Models need to be interpretable and explainable in their work. The initial focus has been on unimodal applications, and now multimodal interpretability and explainability are being investigated [132].

Bias gets introduced through various sources, impacting the fairness of machine learning models [132]. Bias can be of different levels such as sentence-level or embedding level, and sometimes model replicates and amplify bias in the data. Multimodality increases the chances of bias as each modality, besides having heterogeneous nature, can be subject to bias sources. Removal of bias is a challenging task; despite using pre-training on 400 million image-text pairs of data, CLIP [142] still has a social bias in the model. There has been good progress in understanding fairness and bias in unimodal models than multimodal models. Some of the problems and future directions for multimodal interpretability, explainability, fairness, and bias are highlighted in Table 17.

**Table 17**: Summary of challenges and future directions in interpretability and fairness objective

| Problem Area | Problem Description | Future Directions |
|---|---|---|
| Interpretability and Explainability | • Lack of tools to measure multimodal interpretability and explainability like the ones evolved for unimodal applications.<br>• As observed in VL models, there is an inherent bias in multimodal applications towards a certain modality. | • Develop interactive tools to measure multimodal interpretability and explainability, which can take human feedback and correct themselves.<br>• Avoid human bias by having multiple human inputs or setting dynamic and adaptive limits.<br>• Use domain knowledge with graph-based methods to improve interpretability and explainability.<br>• Measure robustness of explainability techniques for co-learning objectives - adversarial perturbations, missing modalities, and noisy conditions. |
| Bias and Fairness | • Definition of fairness score for the multimodal system is not available.<br>• All biases are not well represented; e.g., race and gender are mostly covered in various applications.<br>• The impact of the increase in modalities and implementation of | • Create definitions of fairness score based on bias in algorithms, data, and models for multimodality or an approach to combine unimodal fairness scores.<br>• Have a common framework to test the fairness of multimodal models like |
| | co-learning objectives on bias and fairness is not well understood. | FairCVtest and balanced datasets like FairFace [143] even for missing and noisy modalities scenarios<br>• Consider all possible biases, including those trivial for humans, and test those using co-learning techniques. |

### *7.10 Summary of open problems and future directions for multimodal co-learning*

In this survey, we discussed multimodal deep learning challenges which are pertinent to co-learning and challenges faced by researchers to achieve multimodal co-learning objectives. Lack of common representation, availability of multimodal datasets, and lack of evaluations tools and techniques are three key open problems in multimodal deep learning relevant for co-learning. The key open problems in multimodal co-learning are - all multimodal models should support missing modalities at training and testing fully or partly, need mechanisms to measure the amount of co-learning among the modalities, implementation of co-learning for non-parallel modality data, utilize co-learning to handle label and data noise conditions, reduce dependency on labeled data, availability of multiple domain datasets for domain adaptation, and to ensure interpretability and fairness along with co-learning objectives. These keys challenges are summarized in Table 18.

**Table 18**: Summary of challenges and future directions for multimodal co-learning

| Problem Area | Problem Description | Future Directions |
|---|---|---|
| Representation | Existing multimodal representations are task dependant and limit their use across different tasks and modalities. | Design configurable representation to process varied modalities using pre-trained multimodal models. |
| Datasets | Available datasets are small in sizes, for selected applications and domains, and in the controlled environments only. | Leading researchers and institutes should collaborate to create open multimodal datasets taking advantage of the internet along with synthetic data techniques. |
| Evaluation | Require performance evaluation metrics and tools specific to multimodality and to assess co-learning objectives. | Create performance and robustness measures and report their baselines for major tasks. |
| Missing and Noisy Modalities | Models are not designed to handle real-life conditions of missing and noisy modalities. | Design and test all multimodal models for missing modalities and noise conditions by creating APIs in deep learning library |
| Paired and Labelled data | Dependency on paired and labeled data to achieve co-learning in multimodal applications | The ability to process longer sequences and utilize domain knowledge without annotation is required. |
| Interpretability and fairness | Ensuring interpretability and fairness along with co-learning objectives | Measure if missing and noisy modality impacts interpretability and fairness scores. |

## 8. Conclusion

In this survey, we presented a systematic review of the literature on *multimodal co-learning* for its objectives, taxonomy, current implementation methods, applications, datasets, open problems, and future directions. We have carried out an in-depth and comprehensive survey on multimodal co-learning along with a detailed taxonomy including objectives, categories and sub-categories. Taxonomy includes multimodal co-learning objectives – the presence of modalities, data parallelism, noisy modalities, modality annotations, domain adaptation and interpretability and fairness.

Throughout the paper, we have identified and discussed the representative studies that implemented co-learning objectives and listed modalities involved. We believe that the proposed taxonomy will help researchers to check if all aspects of co-learning are considered and direct their efforts to meet those. Further, we highlight the recent multimodal deep learning methods, which are evolving and can be future directions for multimodal co-learning. We hope this survey will help the audience understand the co-learning objectives, current research progress, and significant background for future research.

## CRediT authorship contribution statement

**Anil Rahate:** Conceptualization, Methodology, Writing- original draft & editing, Visualization, Investigation. **Rahee Walambe:** Conceptualization, Formal analysis, Writing - review & editing. **Sheela Ramanna:** Review and suggestions. **Ketan Kotecha:** Project concept design, Revise and proof the paper, Supervision, Approval

## Declaration of Competing Interest

The authors declare that they have no known competing financial interests or personal relationships that could have appeared to influence the work reported in this paper.

## List of Figures

## List of Tables